\documentclass{article}
\usepackage{ijcai23}

\usepackage{times}
\usepackage{soul}
\usepackage{url}
\usepackage[hidelinks]{hyperref}
\usepackage[utf8]{inputenc}
\usepackage[small]{caption}
\usepackage{graphicx}
\usepackage{amsmath}
\usepackage{amssymb}
\usepackage{amsthm}
\usepackage{xcolor}
\usepackage{booktabs}
\usepackage{algorithm}
\usepackage{algorithmic}
\usepackage[switch]{lineno}
\usepackage{comment}

\newcommand{\eobs}{e^{\textsc{\tiny obs}}} 
\newcommand{\etraj}{e^{\textsc{\tiny traj}}}

\newcommand{\expt}{\mathop{\mathbb{E}}} 

\usepackage{threeparttable} 
\usepackage{subfigure} 
\usepackage{amsfonts}
\usepackage{siunitx}
\usepackage{stfloats}
\usepackage{cancel}

\title{DEIR: Efficient and Robust Exploration through \\Discriminative-Model-Based Episodic Intrinsic Rewards}

\author{
    Shanchuan Wan$^1$\and
    Yujin Tang$^2$\and
    Yingtao Tian$^2$\And
    Tomoyuki Kaneko$^1$
    \affiliations
    $^1$The University of Tokyo\\
    $^2$Google Research, Brain Team
    \emails
    swan@game.c.u-tokyo.ac.jp,
    \{yujintang, alantian\}@google.com,
    kaneko@graco.c.u-tokyo.ac.jp
}

\begin{document}

\maketitle

% Placing full-width figure at the bottom of first page.
% https://tug.org/TUGboat/tb35-3/tb111beet-banner.pdf
% 2 tricks are needed.

% Trick 1 of 2:
\begin{figure}[!b]\setlength{\hfuzz}{1.1\columnwidth}
\begin{minipage}{\textwidth}
\vspace*{-1.4em}
\centering
\includegraphics[width=1.0\textwidth]{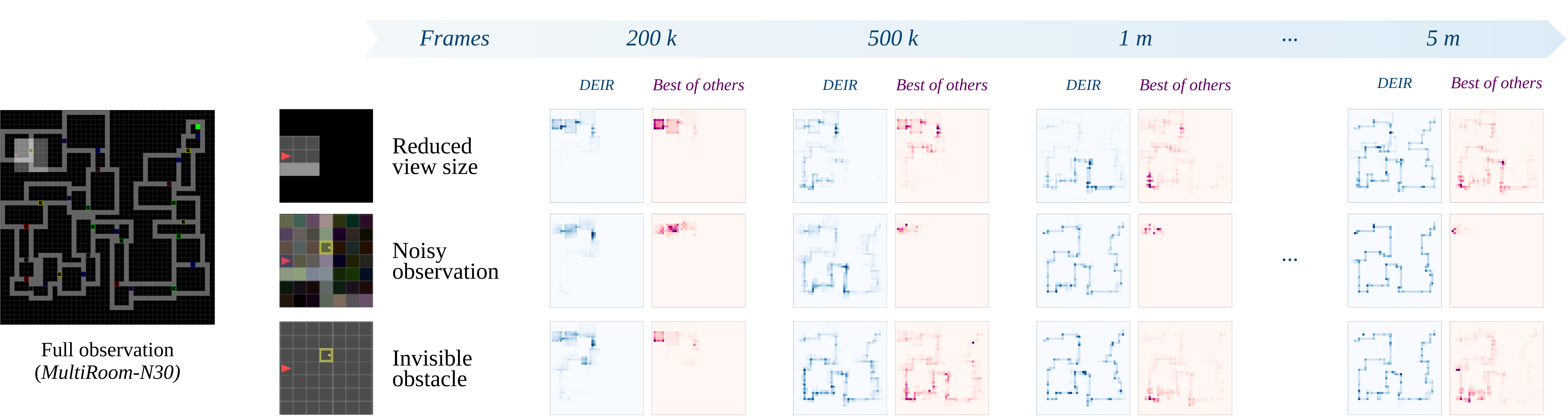}
\caption{
Enlarged \textit{MultiRoom} environment (leftmost figure) from MiniGrid with 30 cascaded rooms as a representative of environments with sparse extrinsic rewards, where the agent (upper-left red dot) is tasked with finding the optimal path to the goal (upper-right green dot). 
We create advanced variants (three rows on the right) with extra difficulties (reduced view sizes, noisy observations, and invisible obstacles). 
In these challenging tasks, existing methods (ICM, RND, NGU, and \text{NovelD}) either fail to find an optimal path or require a prohibitive number of episodes to do so.
In contrast, our proposed DEIR leverages a conditional mutual information-based intrinsic reward and a contrastive learning-inspired model, which is capable of diverse exploration and delivering significantly better performances (see Section~\ref{sec:robustness-exp}). }
\label{figMRN30}
\vspace*{-0.5em}
\end{minipage}
\end{figure}

\begin{abstract}
Exploration is a fundamental aspect of reinforcement learning (RL), and its effectiveness is a deciding factor in the performance of RL algorithms, especially when facing sparse extrinsic rewards. Recent studies have shown the effectiveness of encouraging exploration with intrinsic rewards estimated from novelties in observations. However, there is a gap between the novelty of an observation and an exploration, as both the stochasticity in the environment and the agent's behavior may affect the observation. To evaluate exploratory behaviors accurately, we propose \textbf{DEIR}, a novel method in which we theoretically derive an intrinsic reward with a conditional mutual information term that principally scales with the novelty contributed by agent explorations, and then implement the reward with a discriminative forward model. Extensive experiments on both standard and advanced exploration tasks in MiniGrid show that DEIR quickly learns a better policy than the baselines. Our evaluations on ProcGen demonstrate both the generalization capability and the general applicability of our intrinsic reward. Our source code is available at \linebreak \url{https://github.com/swan-utokyo/deir}.
\end{abstract}

\section{Introduction}

Exploration is an important aspect of reinforcement learning (RL),
as suggested by the famous exploration-exploitation trade-off~\cite{SuttonBarto2018} wherein an agent that only exploits with the current policy would be stuck and fail to improve its policy anymore due to the lack of novel experiences. 
% Trick 2 of 2. 
% >> number here needs manual tuning! <<
% this command need to be placed on a postion of text that is on the  second column, and the argument need to be manually adjusted.
\enlargethispage{-19.0\baselineskip}
Effective exploration is non-trivial, especially in tasks where environmental rewards are sparse.
Relying on unstructured exploration (e.g., $\epsilon$-greedy or randomized probability matching~\cite{SuttonBarto2018,scott2010modern}) requires an exponential number of samples and is unlikely to achieve a satisfactory level of exploration.
Manually designing dense rewards with domain knowledge has exhibited promising results in several areas where RL has significantly progressed, such as game-playing and robotics~\cite{mnih2015human,baker2019emergent,hafner2019dream}. 
However, given the huge amount of knowledge and effort required, designing such dense rewards is only feasible in a handful of tasks, and effective exploration thus remains a challenge.

To tackle this issue, several works have proposed guiding the exploration with intrinsic rewards or rewards that are internal to agents, including ICM~\cite{pathak2017icm}, RND~\cite{burda2019rnd}, NGU~\cite{badia2020ngu}, and \text{NovelD}~\cite{zhang2021noveld}. 
In these methods, the intrinsic reward is devised to encourage visiting states that are likely to be more novel, where novelty is defined as either the distance between the current and past observations or the difference between model predictions and realities.
Although these works have shown encouraging empirical results leading to better exploration efficiency, the relationship between the observed novelty and the agent's actions has not yet been explicitly decoupled.
In other words, effectively handling trivial novelties remains unaddressed, as they are rooted in the stochasticity in the environment's dynamics and have little to do with the agent's exploration capabilities (e.g., the ``noisy TV'' problem~\cite{pathak2017icm}).

This paper bridges this gap with \textbf{DEIR} (\textbf{D}iscriminative-model-based \textbf{E}pisodic \textbf{I}ntrinsic \textbf{R}eward), a novel intrinsic reward design that considers not only the observed novelty but also the effective contribution brought by the agent.
In DEIR, we theoretically derive an intrinsic reward by scaling the novelty metric with a conditional mutual information term between the observation distances and the agent's actions. We make the intrinsic reward tractable by using a simple formula serving as the lower bound of our primary objective.
DEIR is thus designed to distinguish between the contributions to novelty caused by state transitions and by the agent's policy.
For computing the proposed reward, we devise a discriminative forward model that jointly learns the environment's dynamics and the discrimination of genuine and fake trajectories.
We evaluated DEIR in both standard and advanced MiniGrid~\cite{gym_minigrid} tasks (grid-world exploration games with no extrinsic reward until reaching the goal) and found that it outperforms existing methods on both (see Figure~\ref{figMRN30}).
To examine DEIR's generalization capability in tasks with higher dimensional observations, we also conducted experiments in ProcGen~\cite{procgen_github,cobbe2020procgen} (video games with procedurally generated levels that require planning, manipulation, or exploration) and found that it delivers a state-of-the-art (SOTA) performance in all selected tasks.
Finally, we performed an in-depth analysis of our method through additional experiments to clarify the effectiveness of DEIR and its components.

Our contributions can be summarized as follows. (1) Our method, theoretically grounded, effectively decouples the stochasticity in the environment and the novelty gained by the agent's exploration.
(2) Our method empirically outperforms others in this line of work by a large margin, especially in advanced MiniGrid tasks, but not limited thereto. It also has potential applications for a variety of other tasks.
(3) Our method is easy to implement and use, and provides state representations useful for potential downstream objectives. 
\vspace*{-0.3em}

\section{Related Work}
\label{SecRelatedWorkds}

Exploration through intrinsic rewards is widely studied in RL literature, with many works falling into one of two categories:
\emph{Prediction error-driven methods} that are motivated by the differences (``surprises'') between predictions and realities, and
\emph{Novelty-driven methods} that seek novel agent observations.

\textbf{Prediction error-driven methods} generally learn a model of the environment's dynamics and use it to make predictions for future states. Large deviations between the predictions and the realities suggest regimes where the model is insufficiently learned. Intrinsic rewards, positively correlated to prediction errors, encourage the agent to explore more in those states. One of the representative works in this category is ICM~\cite{pathak2017icm}, which jointly trains both the forward and the inverse transition models to capture the environment's dynamics better, but only uses the forward model's prediction errors to generate intrinsic rewards for the agent. 
Prediction error-driven methods require approximating the environment's dynamics with neural networks, which is especially difficult in high-dimensional spaces. Still, as demonstrated by ICM, training with auxiliary tasks seems to be worth the effort.

\textbf{Novelty-driven methods} in recent studies often define a distance metric between observations, and then formulate this distance as intrinsic rewards to encourage more exploration in RL roll-outs. 
Early works include count-based methods, which record how many times distinct states are visited and use the count differences as intrinsic rewards~\cite{bellemare2016unifying,ostrovski2017count,tang2017study}. However, due to their simplicity, count-based methods have difficulty in tasks featuring continuous states or high-dimensional observations. To overcome this problem, neural networks have been introduced to encode the states and observations. 
For example, RND~\cite{burda2019rnd} defines the distance as the difference between the outputs of a parameter-fixed target neural network and a randomly initialized neural network. In this approach, the former network is used to be distilled into the latter one, effectively ``evolving'' a distance metric that adjusts dynamically with the agent's experience.
In \text{NovelD}~\cite{zhang2021noveld}, one of the latest works, RND is applied to evaluate the distances between pairs of observations, in which the boundary between explored and unexplored regions is defined as where the distance is larger than a predefined threshold and large intrinsic rewards are provided when the agent crosses the boundaries. In this way, \text{NovelD} encourages the agent to explore in a manner similar to breadth-first search and has already demonstrated a state-of-the-art (SOTA) performance on many MiniGrid tasks. 
Never-Give-Up (NGU)~\cite{badia2020ngu} introduces the inverse model from ICM in its episodic intrinsic reward generation module. Their final intrinsic reward is based on the Euclidean distances of the K-nearest embeddings of the recently visited states.
In practice, measuring novelty often requires analysis of the distribution of an environment's states. Even so, ``noisy TV''-like problems can still occur, where novelty in the observation space is primarily due to the stochasticity in an environment's dynamics. This prevents the agent from achieving meaningful explorations.

Our method incorporates the advantages of the two categories: while we explicitly encourage our agent to seek novel observations, we also rely on a discriminative model to construct a conditional mutual information term that scales novelty in the observation space, incorporating the model-based prediction task from prediction error-driven methods. 
Unlike conventional novelty-driven methods, our conditional mutual information scaling term effectively eliminates the novelties rooted in the environment's stochasticity other than those brought by the agent's explorations. It has granted us better performances (see Figure~\ref{figMRN30}). 
To illustrate the difference from existing prediction error-driven methods, our model learns both the environment's dynamics and the capability to tell the genuine and fake trajectories apart. Consistent with the results reported in contrastive learning studies~\cite{laskin2020curl,agarwal2021contrastive}, the discriminative nature of our model encodes the observations in a space closely related to the underlying tasks and thus enables us to measure the distances between observations more accurately.

\section{Proposed Method}

\subsection{Background and Notations}

A Markov decision process (MDP)~\cite{SuttonBarto2018} can be defined as a tuple $(S, A, r, f, P_0, \gamma)$, where $S$ is the state space, $A$ is the action space, $r(s_t, a_t, s_{t+1})$ is the reward function, $f(s_t, a_t)$ is the environment's transition function, $P_0$ is the distribution of the initial state $s_0$, and $\gamma \in [0, 1]$ is the reward discount factor. The goal is to optimize a policy $\pi: S \times A \to \mathbb{R}$ so that the expected accumulated reward $\mathbb{E}_{s_0 \sim P_0}[\sum_t {\gamma^t r(s_t, a_t, s_{t+1}})]$ is maximized. 
However, in a partially observable MDP (POMDP), $s_t \in S$ is not accessible to the agent. 
A common practice is to use $\pi(a_t|\tau_t)$ for the policy instead, where $\tau_t=\{o_0, o_1, \cdots, o_t\}$ is an approximation of $s_t$. Many works implement this using recurrent neural networks (RNN)~\cite{rumelhart1986learning} to best utilize available historical information.

% Per Reviewer #1 Q4
We adopt proximal policy optimization (PPO)~\cite{schulman17ppo} to learn the agent's policy.
With PPO as a basis, we propose an episodic intrinsic reward that helps decouple the novelties introduced by the agent from those by the environment. We also introduce a discriminative forward model to learn better state representations in partially observable environments.
Following popular studies, our reward function is designed to be the weighted sum of extrinsic (those from the environment) and intrinsic (those from curiosity) rewards: $r(s_t, a_t, s_{t+1})=r_t^{\textsc{\tiny E}} + \beta \cdot r_t^{\textsc{\tiny I}}$, where $r_t^{\textsc{\tiny E}}$ and $r_t^{\textsc{\tiny I}}$ (both functions of $(s_t, a_t, s_{t+1})$) are respectively the extrinsic and intrinsic rewards at time step $t$, and $\beta$ is a hyperparameter. 
    
\subsection{Episodic Intrinsic Reward}
\label{sec:episodic-intrinsic-reward}

\subsubsection{Scaling the Novelty}
For a pair of states $(s_t, s_i), \forall i\in[0, t)$ and the action $a_t$, 
we want a novel state (valuable for exploration) to have a large distance between observations $(o_{t+1}, o_i)$ while that distance is closely related to the action $a_t$.
Intuitively, it is crucial to distinguish novelty rooted in the environment's stochasticity from novelty brought by an exploratory action of the agent. This leads to our primary objective, which is to maximize 
\begin{equation}
J=\mathrm{dist}(o_{t+1}, o_i) \cdot I \left( \mathrm{dist}(o_{t+1}, o_i) ; a_t | s_t, s_i \right)
\label{eqOriginalTargetMainText}
\end{equation}
as a product of the distance $\mathrm{dist}(o_{t+1}, o_i)$ (denoted as $D_{t+1,i}$) between observations and conditional mutual information
\begin{multline*}
I(D_{t+1,i}; a_t | s_t, s_i) = \\
      \expt_{s_t, s_i, a_t} [ \mathrm{D_{KL}}(p(D_{t+1,i} | s_t, s_i, a_t) \| p(D_{t+1,i} | s_t, s_i))].
\end{multline*}

With the Bretagnolle–Huber inequality~\cite{bretagnolle1978estimation}, we have 
$$
\mathrm{D_{KL}}(P \| Q) \geq - \mathrm{log} (1 - \mathrm{d^2_{TV}}(P, Q) ),
$$
where $P(x) = p(x | s_t, s_i, a_t)$ and $Q(x) = p(x | s_t, s_i)$ \pagebreak are defined for simplicity, and $\mathrm{d_{TV}}(P, Q)=\frac{1}{2}\|P - Q\|_1 $ is the total variation between $P$ and $Q$. 

Note that, in deterministic environments (including partially observable cases), 
(a) $P$ is a unit impulse function that has the only non-zero value at $D_{t+1,i}$, and (b) we can naturally assume the  $D_{t+1,i}|s_t,s_i \sim \mathrm{Exp}({{\lambda}= {1} / {\mathrm{dist}(s_t, s_i)} })$ to match the distance between observations and that of their underlying states.
Thus, we devise a simplified surrogate function for the mutual information, as
\[
\mathrm{D_{KL}}(P \| Q) \geq \mathrm{log}{\left( \mathrm{dist}(s_t, s_i) \right)} + \frac{\mathrm{dist}(o_{t+1}, o_i)}{\mathrm{dist}(s_t, s_i)} + \mathrm{const}.
\]

Substituting it back to the objective, we obtain
\begin{multline}
    J \ge \mathrm{dist}(o_{t+1}, o_i) \\
    \cdot \expt_{s_t, s_i, a_t} \left( \mathrm{log}{\left( \mathrm{dist}(s_t, s_i) \right)} + \frac{\mathrm{dist}(o_{t+1}, o_i)}{\mathrm{dist}(s_t, s_i)} \right).
    \label{EqExpectationTarget}
\end{multline}

To make it tractable, we simplify the right-hand side to
\begin{equation}
  J \ge \min_{i}{\frac{\mathrm{dist}^2(o_{t+1}, o_i)}{\mathrm{dist}(s_t, s_i)}} ,
  \label{EqProposedTarget}
\end{equation}
which is a lower bound for the original objective.
It is simpler and empirically performs as well as or even better than Equation~\ref{EqExpectationTarget}.
We speculate that improving the minimum value is crucial to improving the expectation value. A detailed derivation is provided in Appendix~\ref{sec:intrinsic-reward-derivation}.

\subsubsection{Intrinsic Reward Design}

To maximize our objective through the lower bound in Equation~\ref{EqProposedTarget}, we propose our intrinsic reward in an episodic manner:

\vspace*{-0.4em}
\begin{equation}
r_t^{\textsc{\tiny I}} = \min_{\forall i \in \left[0, t\right)}
\left\lbrace 
\frac{ \mathrm{dist}^2 (\eobs_i, \eobs_{t+1}) }{ \mathrm{dist}(\etraj_i, \etraj_t) + \epsilon}
\right\rbrace,
\label{EqProposedIntrinsicReward}
\end{equation}
\vspace*{0.1em}

\noindent where $\eobs_t$ is the embedding of observation $o_t$, $\etraj_t$ is the embedding of trajectory $\tau_t$ at time step $t$ in an episode, $\mathrm{dist}$ is the Euclidean distance between two embeddings, and $\epsilon$ is a small constant ($10^{-6}$ in our experiments) for numeric stability. 
Note that the complete information regarding $s_t$ is inaccessible in POMDPs, so $\etraj_t$ is commonly used as a proxy of $s_t$. 
All these constitute a metric space of observations where distance is defined. 
Both $\eobs_t$ and $\etraj_t$ are computed by a discriminative forward model, as detailed in Section~\ref{SubSecDiscriminativeModel}.

Our intrinsic rewards are episodic, as they are created from the observations seen in a single episode. 
Arguably, episodic rewards are generally compatible with lifelong rewards and can be used jointly. Still, in this work, we focus on the former for simplicity and leave such a combination for future studies.

\subsection{Learning a Discriminative Model}
\label{SubSecDiscriminativeModel}

\begin{figure}[!t]
    \centering
    \includegraphics[width=1\columnwidth]{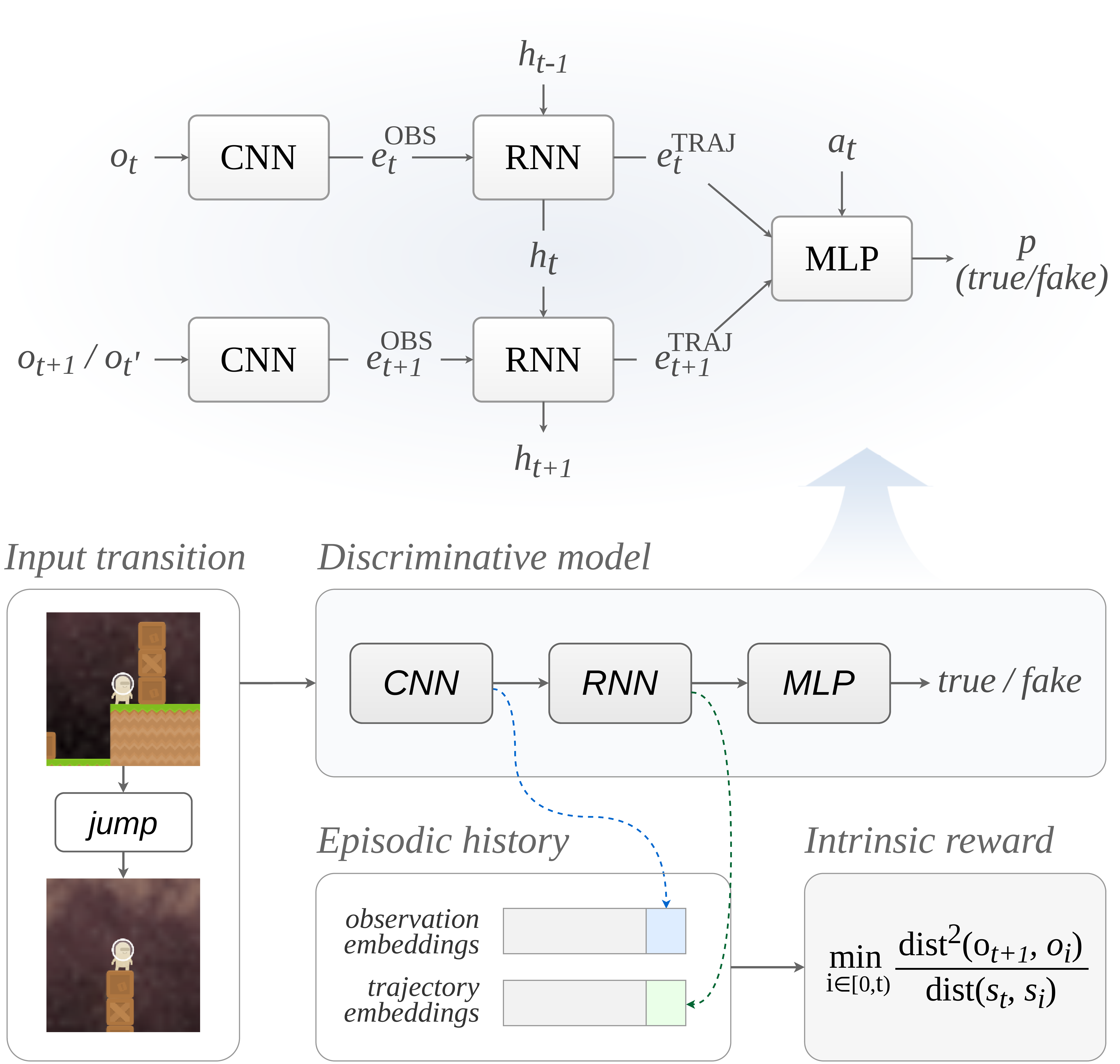}
    \caption{
    Overview of DEIR.
    Given the \emph{input transition} of two observations and an action,
    the \emph{discriminative model} predicts whether they are from a truly observed transition.
    Observation and trajectory embeddings produced by the model are saved in an \emph{episodic history}
    to compute \emph{intrinsic rewards} for guiding explorations.
    }
    \label{figArch}
    \vspace*{-1em}
\end{figure}

We train a neural network to extract embeddings $\eobs_t$ and $\etraj_t$ from observations in high-dimensional spaces. 
Existing studies have adopted auxiliary tasks in which the forward and inverse models are two representatives to obtain better embeddings suitable for exploration. We propose an improved auxiliary task suitable for exploration in POMDPs inspired by contrastive learning~\cite{laskin2020curl}. 
Concretely, our proposed model learns the environment's dynamics and discriminates the genuine trajectories from the fake. 
Figure~\ref{figArch} illustrates the architecture of our model.

\subsubsection{Model Definition}
\label{sec:model-definition}

Our discriminative model is denoted as $\mathrm{Dsc}(o_t, a_t, o_x): O \times A \times O' \to [0, 1]$, where $o_t, a_t$ are defined as above and $o_x \in \{o_{t+1}, o_{t'}\}$ is either the next observation $o_{t+1}$ (positive example) or an observation $o_{t'}$ that has been observed recently and is randomly selected at time step $t$ (negative example) for each training sample. In short, the discriminative model estimates the likelihood that the input $o_x$ is a positive example.
Particularly, to efficiently retrieve fake examples with decent diversity during RL roll-outs, we maintain recent novel observations with a first-in-first-out queue $\mathbb{Q}$ so that $o_{t'}$ can be randomly selected from all observations saved in $\mathbb{Q}$. 
This queue is maintained by continuously adding newly retrieved ``novel'' observations to replace the oldest observations. Here, $o_t$ is considered ``novel'' only if $r^I_t$ is not less than the running average of all intrinsic rewards. 
In addition, when sampling fake observations from $\mathbb{Q}$, we always sample twice and keep only ones that differ from the true ones.
These measures lead to a very high ratio of valid samples with distinct true and fake observations in training.
Algorithms used in modeling are further detailed in Appendix~\ref{sec:algorithms-used-in-modeling}.

\subsubsection{Proposed Architecture}
        
The proposed intrinsic reward is based on the observation embeddings $e_t^{\textsc{\tiny obs}}$ and the trajectory embeddings $e_t^{\textsc{\tiny traj}}$ generated by the discriminator. 
From input to output, the discriminator $\mathrm{Dsc}$ is formed by a convolutional neural network (CNN)~\cite{lecun1998gradient}, a recurrent neural network (RNN)~\cite{rumelhart1986learning}, and a multi-layer perceptron (MLP)~\cite{rosenblatt1958perceptron} output head. We adopt gated recurrent units (GRU)~\cite{cho2014gru} in the RNN module to reduce the number of trainable parameters (compared to LSTM~\cite{hochreiter1997long}).
As shown in Figure~\ref{figArch}, the CNN takes the observations $o_t$ and $o_{t+1}$ as input in parallel and outputs two observation embeddings. These observation embeddings are then fed into the RNN, together with the RNN's previous hidden state $h_{t-1}$. In addition to the updated hidden state $h_t$, the RNN outputs the embeddings of two trajectories starting from the beginning of an episode and ending at time steps $t$ and $t+1$, respectively. 
Finally, the two trajectory embeddings with the action $a_t$ are fed into the MLP for predicting the likelihood. 
RNN hidden states are saved as part of the discriminative model's training samples.
We adopt PPO for learning the policy, as it refreshes the experience buffer more frequently than other algorithms. 
Therefore, we do not apply specialized methods to renew the hidden states within one episode.

\subsubsection{Mini-Batches and Loss Function}

During training, each mini-batch consists of two types of samples: half of them positive and half of them negative. Both types are picked from the agent's recent experiences. The discriminator is trained with a binary cross-entropy loss function \cite{pml1Book} as in ordinary classification tasks. 

\section{Experiments}

We address the following questions through our experiments:
\begin{itemize}
    \item Is DEIR effective in standard benchmark tasks and can it maintain its performance in more challenging settings?
    \vspace*{-0.2em}
    \item Is our design decision in DEIR generally applicable to a variety of tasks, and particularly, can it generalize to tasks with higher dimensional observations?
    \vspace*{-0.2em}
    \item How significantly does each technical component in DEIR contribute to the performance?
    \vspace*{-0.2em}
\end{itemize}

\begin{figure*}[!p]
\centering
\includegraphics[width=0.96\textwidth]{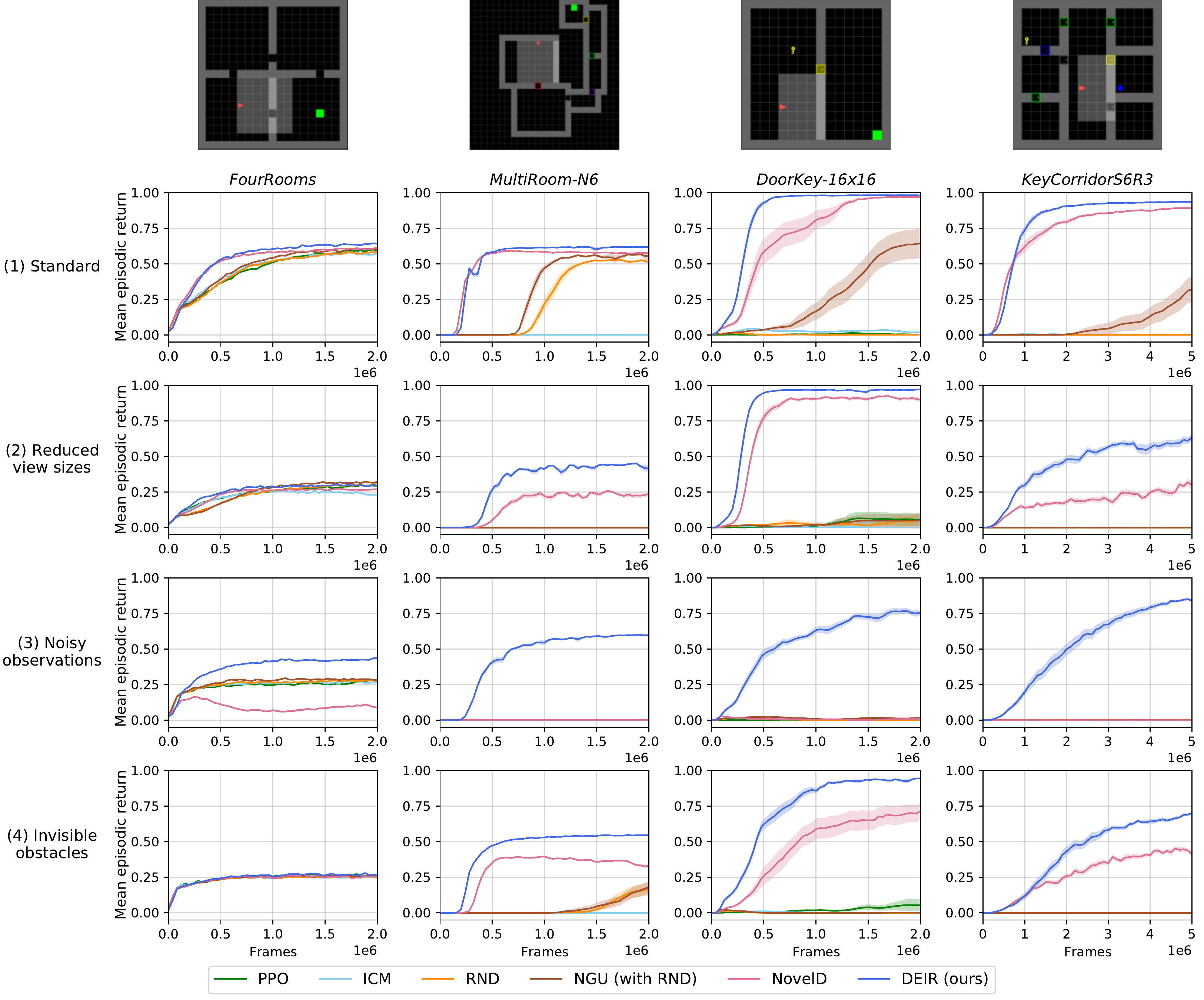}
\vspace{-0.3em}
\caption{
Mean episodic returns in (1) standard and (2)--(4) advanced MiniGrid games.
(1) Agent has a fixed $7 \times 7$, unhindered view size.
(2) Agent has a reduced view size of $3 \times 3$ grids. 
(3) Noisy observations, where Gaussian noise ($\mu = 0.0, \sigma = 0.1$) are added element-wise to the observations that are first normalized to $[0,1]$. 
(4) Obstacles that are invisible to the agent but still in effect. 
In all figures, Y axes start at $-0.05$ to show near-zero values.
}
\label{figMinigrid}
\end{figure*}

\begin{figure*}[!p]
\centering
\includegraphics[width=0.96\textwidth]{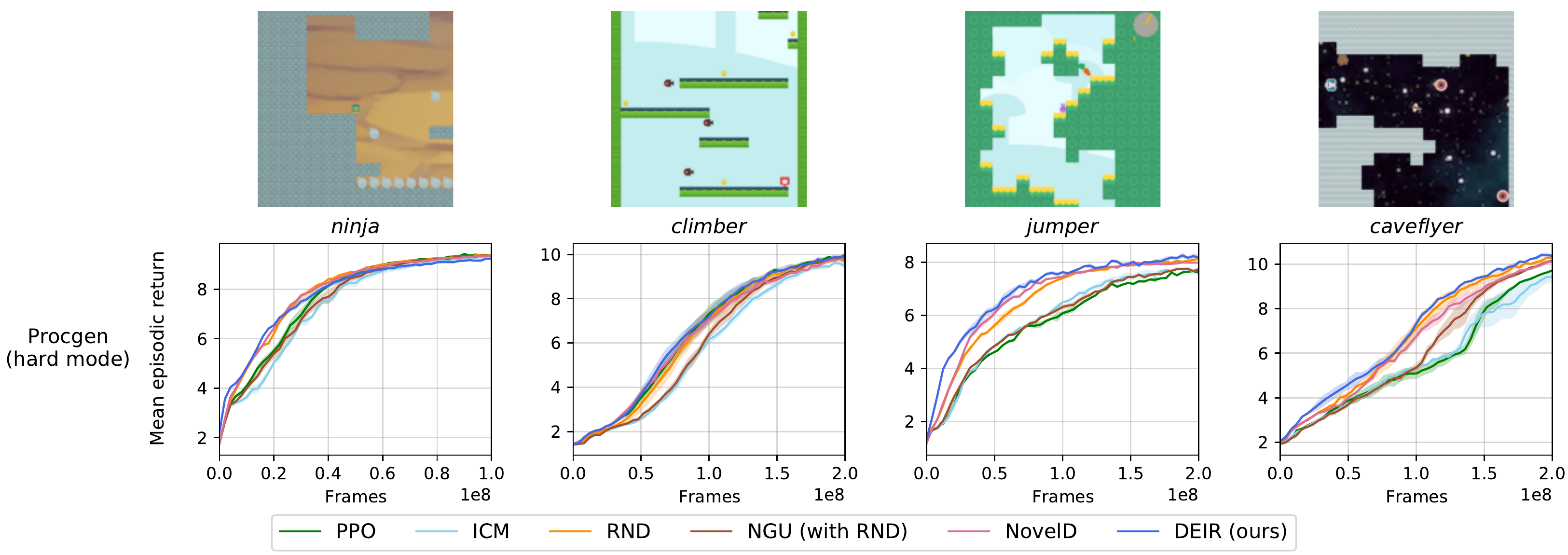}
\vspace{-0.3em}
\caption{
Mean episodic returns in ProcGen (hard mode) games. Episodes are randomly generated from the full distribution of game levels.
}
\label{figProcGen}
\end{figure*}

\subsection{Experimental Setup}
\label{sec:experimental-setup}

We evaluate DEIR using the following two popular procedurally generated RL benchmarks. 
(1)~\textbf{MiniGrid}~\cite{gym_minigrid}, which consists of 20 grid-world exploration games featuring different room layouts, interactive objects, and goals. An agent needs to learn a specific sequence of actions to reach a final goal with its limited view size. Valid actions include picking up a key, unlocking a door, unpacking a box, and moving an object. No extrinsic reward is given until the goal.
(2)~\textbf{ProcGen}~\cite{cobbe2020procgen}, which consists of 16 games with $64 \times 64 \times 3$ RGB image inputs, each of which requires a certain level of planning, manipulation, or exploration skill to pass. Each episode is a unique game level with randomly initialized map settings, physical properties, enemy units, and visual objects. The agent needs to learn policies that can be generalized to unseen levels.

Environments and networks are by default initialized with 20 seeds in each MiniGrid experiment and three seeds (from the full distribution of game levels) in each ProcGen experiment. All experimental results are reported with the average episodic return of all runs with standard errors.
We performed hyperparameter searches for every method involved in our experiments to ensure they have the best performance possible.
We also performed sensitivity analyses on two key hyperparameters of our method, namely, the maximum episode length and the maximum observation queue size, and found that both work well with a wide range of candidate values.
Note that, while our experiments in this paper focus on POMDPs, we find that the performance of DEIR is superior to existing exploration methods in many fully observable tasks as well, which deserves further analysis in future works.
We further detailed benchmarks, hyperparameters, networks, and computing environments in Appendix~\ref{sec:benchmark-environments} to \ref{sec:computing-environments} respectively.

\subsection{Evaluation Experiments}
\label{sec:evaluation-experiments}

\subsubsection{Evaluation in Standard MiniGrid}

We first evaluated the performance of DEIR in four standard MiniGrid tasks, where the agent's view size is a $7 \times 7$ square and no other constraints are applied. The mean episodic returns of all exploration methods are shown in Figure~\ref{figMinigrid} (first row).
In \textit{FourRooms} and \textit{MultiRoom-N6}, which are simple tasks, DEIR and the existing methods all exhibited a decent performance. 
\textit{DoorKey-16x16} and \textit{KeyCorridorS6R3} are more complex tasks featuring multiple sub-tasks that must be completed in a particular order, requiring efficient exploration. 
In these complex tasks, DEIR also learned better policies faster than the existing methods.

\begin{figure}[!t]
\centering
\includegraphics[width=1\columnwidth]{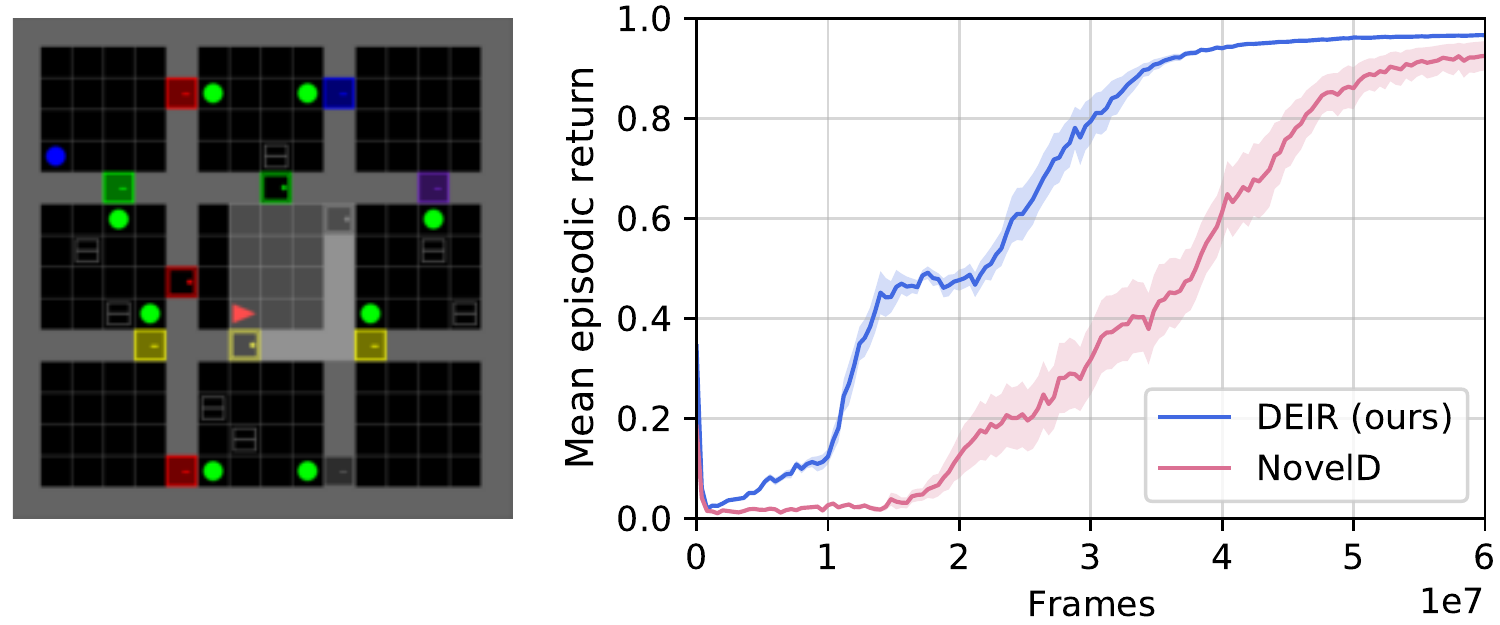}
\caption{
Comparison of episodic returns of DEIR and \text{NovelD} in \textit{ObstructedMaze-Full}---the hardest standard task in MiniGrid that has been solved by only few methods, including \text{NovelD}. DEIR was able to solve the task here and also learned significantly faster.
}
\vspace*{-0.3em}
\label{figOMFull}
\end{figure}

We also compared the performances of DEIR and \text{NovelD} on the most difficult MiniGrid task \textit{ObstructedMaze-Full} (see Figure~\ref{figOMFull}).
Its difficulty is due to the highest number of sub-tasks that need to be completed in the correct order among all standard MiniGrid tasks. So far, it has been solved by only few methods, including \text{NovelD}, albeit in an excessive amount of time. 
To reach the same SOTA performance, DEIR required only around $70\%$ of frames as \text{NovelD}, for which we also conducted hyperparameter searches and exceeded its original implementation by requiring fewer training steps.

\subsubsection{Evaluation in Advanced MiniGrid}
\label{sec:robustness-exp}
We further evaluated the robustness of DEIR on 12 MiniGrid tasks with advanced (more challenging) environmental settings, in which the following modifications were made to the standard environments (see examples in Figure~\ref{figMRN30}).
\begin{description}
\item[Reduced view sizes.] The agent's view size is reduced from the default $7 \times 7$ grids to the minimum possible view size of $3 \times 3$ (81.6\% reduction in area).  Consequently, the agent needs to utilize its observation history effectively.

\item[Noisy observations.] At each time step, noises are sampled from a Gaussian distribution ($\mu = 0.0, \sigma = 0.1$) in an element-wise manner and added to the observations that are first normalized to $[0,1]$.
With this change, each observation looks novel even if the agent does not explore.

\item[Invisible obstacles.] Obstacles are invisible to the agent but still in effect; that is, the agent simply perceives them the same way as floors, but cannot step on or pass through them. 
This requires the agent to have a comprehensive understanding of the environment's dynamics, beyond the superficial observable novelty.

\end{description}

The results in Figure~\ref{figMinigrid} (second to fourth rows) clearly show that DEIR was significantly more robust than the existing methods in all advanced settings.
The performance difference was especially large when only noisy observations were presented.
Compared to the results in standard tasks, DEIR lost at most 25\% of its returns by the end of the training, while other methods lost nearly all of their returns in 75\% of the noisy-observation tasks.
Note that our noisy-observation task is similar to the ``noisy-TV'' experiments~\cite{burda2019rnd} but features increased difficulty due to changing more pixels. 
In ICM~\cite{pathak2017icm}, up to 40\% of the observation image was replaced with white noise.
According to published source code, \text{NovelD} was originally tested in a scenario where at most one special object in an observation switches its color among six hard-coded colors when triggered by the agent, i.e., most parts of the observation image remain unchanged.
Detailed results in MiniGrid are described in Appendix~\ref{sec:detailed-experimental-results-in-minigrid}.

\subsubsection{Generalization Evaluation in ProcGen}
We evaluated the generalization capability of DEIR under various reward settings and environmental dynamics using the ProcGen benchmark. 
Four tasks (on hard mode) were selected on the basis of their partial observability and demand for advanced exploration skills.
We used the same CNN structure as in IMPALA~\cite{espeholt2018impala} and Cobbe et al.'s work~\cite{cobbe2020procgen} for the agent's policy and value function, and a widened version of the CNN used in DQN~\cite{mnih2015human} for the dynamics model of each exploration method. 
The results in Figure~\ref{figProcGen} show that DEIR performed better than or as well as other exploration methods and could successfully generalize to new game levels generated during training. 
The results also suggest that the proposed model and intrinsic reward are universally applicable to a variety of tasks, including those with higher-dimensional observations. 
In addition, we confirmed that the performance of DEIR was consistent with the training results reported in previous studies~\cite{cobbe2020procgen,cobbe2021phasic,raileanu2021decoupling}.

\begin{figure}[!htb]
\centering
\includegraphics[width=1\columnwidth]{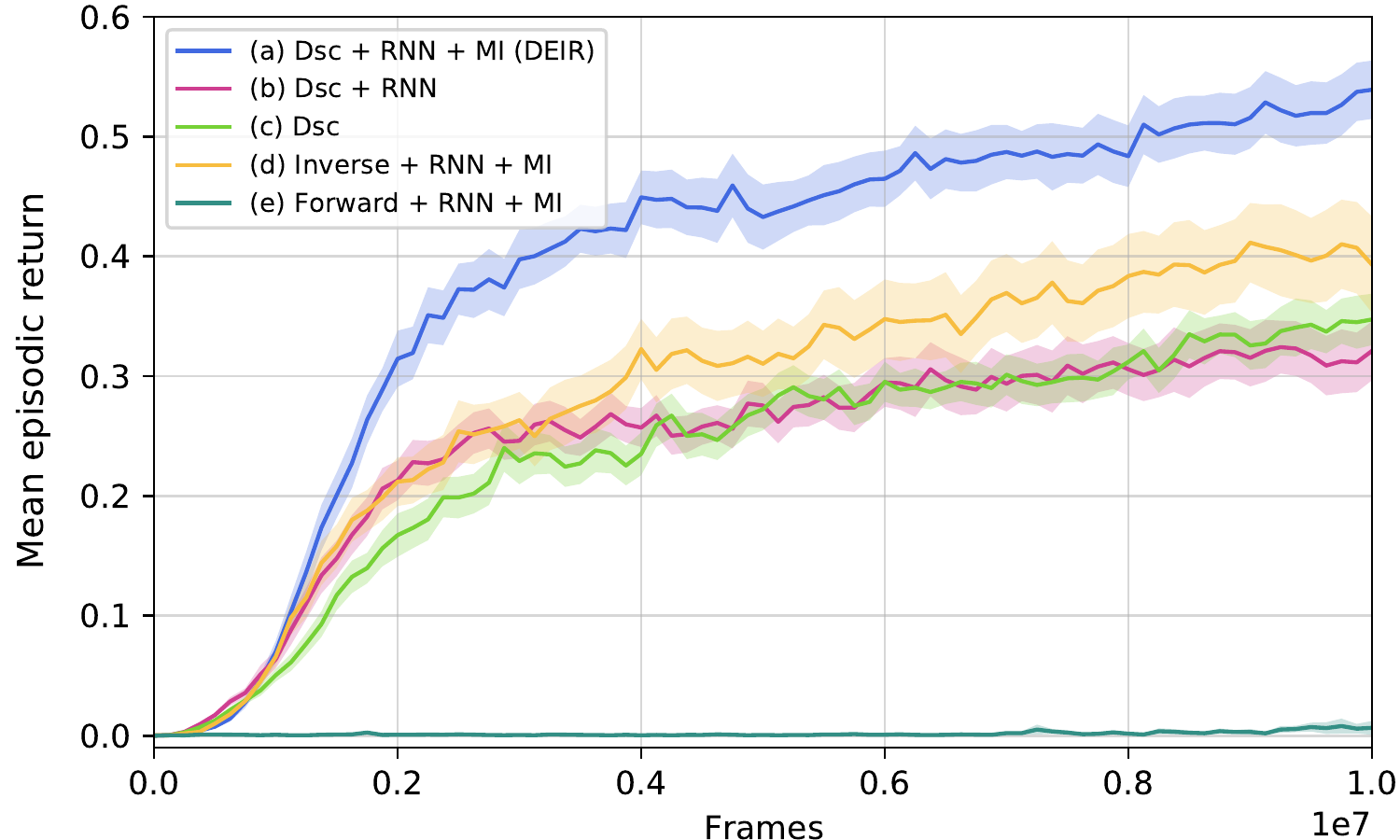}
\caption{
    Results of ablation studies in an advanced \textit{KeyCorridorS6R3} task.
    The effectiveness of the conditional mutual information (MI) can be confirmed by comparing (a), (b), (c). 
    The difference between (a), (d), (e) shows the discriminator's importance.
}
\vspace*{-0.3em}
\label{figAblations}
\end{figure}

\subsection{Ablation Studies}
\label{sec:ablation-studies}

To better understand DEIR, we created an advanced \textit{KeyCorridorS6R3} task with all three modifications proposed in Section~\ref{sec:robustness-exp} (view size: 3, standard deviation of noise: 0.3, obstacles: invisible), and conducted ablation studies to analyze the importance of 
(1) the conditional mutual information term and 
(2) the discriminative model.

\subsubsection{Conditional Mutual Information Scaling}
To analyze the importance of the conditional mutual information term proposed in Equation~\ref{eqOriginalTargetMainText}, we evaluated the performances of our DEIR agent and a variant without the mutual information term (see Figure~\ref{figAblations}(a) and (b)).
As we can see, DEIR performed significantly better than the latter, which demonstrates the importance of our intrinsic reward design.

We utilize RNN in our model to capture temporal information in trajectories, which enables a more accurate representation to be learned. 
Thus, we further examine the effect of RNN on its own by training a separate agent with the discriminator only (see Figure~\ref{figAblations}(c)) and found that using RNN alone barely brings any benefit compared to Figure~\ref{figAblations}(b). 
Thus, we are confident that the mutual information scaling term indeed contributes to all of the performance improvements.

\subsubsection{Discriminative Model}

The performances of the DEIR agents driven by the inverse model and the forward model are shown in Figure~\ref{figAblations}(d) and (e), respectively. 
We applied the conditional mutual information term and RNN to both, and used the same tuned hyperparameters as in our previous experiments. 
Compared with Figure~\ref{figAblations}(a), our discriminative model-driven agent presented an evident advantage over the agent trained with the inverse model alone, while the forward model-driven agent completely failed to learn any meaningful policy in the task (due to the fact that the forward model is notorious for being weak to noise~\cite{pathak2017icm}), suggesting that to achieve an advanced performance, the discriminative model is indispensable for learning state representations in POMDPs. We believe the main contribution stems from the trajectory embeddings learned by the discriminative model, and we present further analysis results in Section~\ref{sec:analysis-on-learned-embeddings} for more insights.

\begin{figure}[!t]
\centering
\includegraphics[width=0.97\columnwidth]{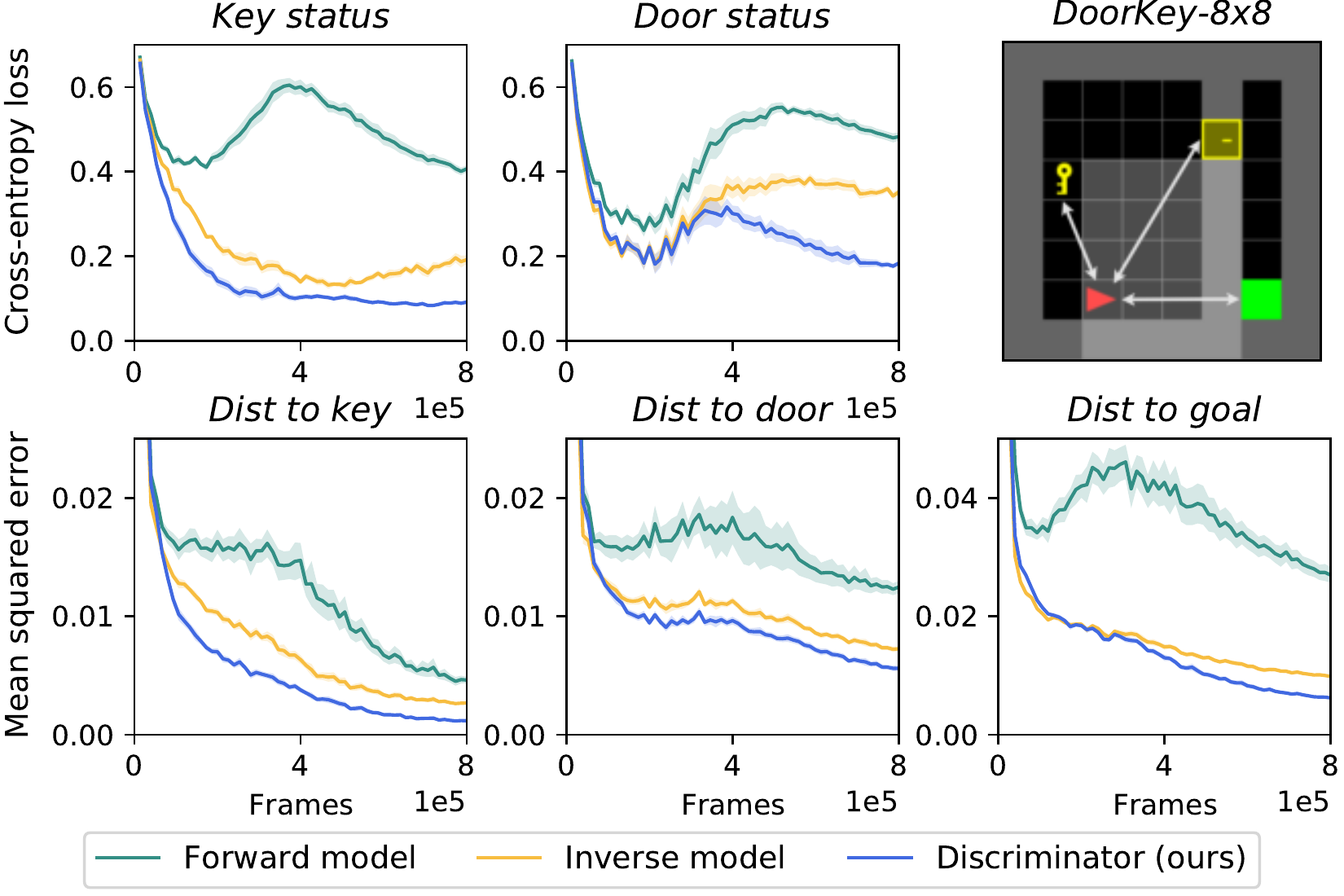}
\caption{
Validation losses of predicting temporal and spatial metrics using trajectory embeddings by three models. 
Upper-right figure exemplifies the objects (key, door, goal) related to the metrics. 
Upper: Whether the agent has picked up the key and opened the door. 
Lower: The agent's normalized distances to the key, door, and goal.
}
\vspace*{-0.3em}
\label{figStatus}
\end{figure}

\subsection{Effectiveness of Learned Embeddings}
\label{sec:analysis-on-learned-embeddings}
To obtain further insights into the impact of the learned embeddings on the final performance, we conducted an experiment to compare three model variants. Concretely, we trained a forward-only, an inverse-only, and a discriminative model using the same data sampled by a vanilla PPO agent in standard \textit{DoorKey-8x8}, where an agent needs to find a key to open a door that leads to the goal in another room.
Following Alain and Bengio's work~\cite{alain2016understanding}, we devise five auxiliary supervised learning tasks. Given the learned trajectory embeddings $\etraj$, each task predicts an important temporal or spatial metric that is directly related to the game's progress or the agent's position.
Technically, the ground-truth metrics are retrieved from the game's backend, and training stops after the agent reaches a near-optimal policy. 
The results in Figure~\ref{figStatus} demonstrate that our discriminator-based method learns the most helpful embeddings for predicting important temporal and spatial metrics. This explains why it can benefit downstream objectives in RL, including exploration.

In comparison, embeddings from the forward and inverse models did not perform as well as ours, which is consistent with the findings in Figure~\ref{figAblations}. 
We hypothesize that the inverse model is less motivated to include historical data in embeddings because, by design, it can reliably infer actions merely from its always-factual inputs.
On the other hand, the forward model relies too much on visual details, so its embeddings do not carry much information crucial to underlying tasks~\cite{gulrajani2017pixelvae,alias2017zforing}.

\vspace*{-0.2em}
\section{Conclusion}

Training RL agents to explore effectively is a challenging problem, especially in environments with sparse rewards. 
A promising approach is to augment the extrinsic rewards with novelty-driven intrinsic rewards. 
However, focusing only on the novelty of observations is insufficient because an agent may incorrectly recognize the stochasticity in the environment's dynamics as novelties brought by its explorations.
In this work, we proposed scaling the observation novelty with a conditional mutual information term that explicitly relates the agent's actions to the distances between observations, and learning a discriminative model that gives better intrinsic rewards.
Compared with baselines, our method delivers outstanding performances in both standard and advanced versions of MiniGrid tasks. Also, it demonstrates general applicability to a variety of tasks with higher-dimensional inputs (such as those in ProcGen). 
As future work, we envision research on continuous action spaces and multi-agent settings.

\bibliographystyle{named}
\bibliography{ijcai23}

\begin{thebibliography}{}

\bibitem[\protect\citeauthoryear{Agarwal \bgroup \em et al.\egroup
  }{2021}]{agarwal2021contrastive}
Rishabh Agarwal, Marlos~C. Machado, Pablo~Samuel Castro, and Marc~G. Bellemare.
\newblock Contrastive behavioral similarity embeddings for generalization in
  reinforcement learning.
\newblock {\em arXiv preprint arXiv:2101.05265}, 2021.

\bibitem[\protect\citeauthoryear{Alain and
  Bengio}{2016}]{alain2016understanding}
Guillaume Alain and Yoshua Bengio.
\newblock Understanding intermediate layers using linear classifier probes.
\newblock {\em arXiv preprint arXiv:1610.01644}, 2016.

\bibitem[\protect\citeauthoryear{Andrychowicz \bgroup \em et al.\egroup
  }{2021}]{andrychowicz2021matters}
Marcin Andrychowicz, Anton Raichuk, Piotr Sta{\'n}czyk, Manu Orsini, Sertan
  Girgin, Rapha{\"e}l Marinier, L{\'e}onard Hussenot, Matthieu Geist, Olivier
  Pietquin, Marcin Michalski, et~al.
\newblock What matters in on-policy reinforcement learning? a large-scale
  empirical study.
\newblock In {\em International Conference on Learning Representations}, 2021.

\bibitem[\protect\citeauthoryear{Ba \bgroup \em et al.\egroup
  }{2016}]{ba2016layer}
Lei~Jimmy Ba, Jamie~Ryan Kiros, and Geoffrey~E. Hinton.
\newblock Layer normalization.
\newblock {\em CoRR}, abs/1607.06450, 2016.

\bibitem[\protect\citeauthoryear{Badia \bgroup \em et al.\egroup
  }{2020}]{badia2020ngu}
Adri{\`a}~Puigdom{\`e}nech Badia, Pablo Sprechmann, Alex Vitvitskyi, Daniel
  Guo, Bilal Piot, Steven Kapturowski, Olivier Tieleman, Martin Arjovsky,
  Alexander Pritzel, Andrew Bolt, et~al.
\newblock Never give up: Learning directed exploration strategies.
\newblock In {\em International Conference on Learning Representations}, 2020.

\bibitem[\protect\citeauthoryear{Baker \bgroup \em et al.\egroup
  }{2019}]{baker2019emergent}
Bowen Baker, Ingmar Kanitscheider, Todor Markov, Yi~Wu, Glenn Powell, Bob
  McGrew, and Igor Mordatch.
\newblock Emergent tool use from multi-agent autocurricula.
\newblock In {\em International Conference on Learning Representations}, 2019.

\bibitem[\protect\citeauthoryear{Bellemare \bgroup \em et al.\egroup
  }{2016}]{bellemare2016unifying}
Marc Bellemare, Sriram Srinivasan, Georg Ostrovski, Tom Schaul, David Saxton,
  and Remi Munos.
\newblock Unifying count-based exploration and intrinsic motivation.
\newblock {\em Advances in neural information processing systems}, 29, 2016.

\bibitem[\protect\citeauthoryear{Blundell \bgroup \em et al.\egroup
  }{2016}]{blundell2016model}
Charles Blundell, Benigno Uria, Alexander Pritzel, Yazhe Li, Avraham Ruderman,
  Joel~Z. Leibo, Jack Rae, Daan Wierstra, and Demis Hassabis.
\newblock Model-free episodic control.
\newblock {\em arXiv preprint arXiv:1606.04460}, 2016.

\bibitem[\protect\citeauthoryear{Bretagnolle and
  Huber}{1978}]{bretagnolle1978estimation}
Jean Bretagnolle and Catherine Huber.
\newblock Estimation des densit{\'e}s: risque minimax.
\newblock {\em S{\'e}minaire de probabilit{\'e}s de Strasbourg}, 12:342--363,
  1978.

\bibitem[\protect\citeauthoryear{Brockman \bgroup \em et al.\egroup
  }{2016}]{openai_gym}
Greg Brockman, Vicki Cheung, Ludwig Pettersson, Jonas Schneider, John Schulman,
  Jie Tang, and Wojciech Zaremba.
\newblock Openai gym.
\newblock {\em CoRR}, abs/1606.01540, 2016.

\bibitem[\protect\citeauthoryear{Burda \bgroup \em et al.\egroup
  }{2019}]{burda2019rnd}
Yuri Burda, Harrison Edwards, Amos Storkey, and Oleg Klimov.
\newblock Exploration by random network distillation.
\newblock In {\em Seventh International Conference on Learning
  Representations}, 2019.

\bibitem[\protect\citeauthoryear{Chevalier-Boisvert \bgroup \em et al.\egroup
  }{2018}]{gym_minigrid}
Maxime Chevalier-Boisvert, Lucas Willems, and Suman Pal.
\newblock Minimalistic gridworld environment for openai gym.
\newblock \url{https://github.com/maximecb/gym-minigrid}, 2018.

\bibitem[\protect\citeauthoryear{Cho \bgroup \em et al.\egroup
  }{2014}]{cho2014gru}
Kyunghyun Cho, Bart van Merri{\"e}nboer, Dzmitry Bahdanau, and Yoshua Bengio.
\newblock On the properties of neural machine translation: Encoder--decoder
  approaches.
\newblock {\em Syntax, Semantics and Structure in Statistical Translation},
  page 103, 2014.

\bibitem[\protect\citeauthoryear{Cobbe \bgroup \em et al.\egroup
  }{2019}]{procgen_github}
Karl Cobbe, Christopher Hesse, Jacob Hilton, and John Schulman.
\newblock Procgen benchmark.
\newblock \url{https://github.com/openai/procgen}, 2019.

\bibitem[\protect\citeauthoryear{Cobbe \bgroup \em et al.\egroup
  }{2020}]{cobbe2020procgen}
Karl Cobbe, Chris Hesse, Jacob Hilton, and John Schulman.
\newblock Leveraging procedural generation to benchmark reinforcement learning.
\newblock In {\em International Conference on Machine Learning}, pages
  2048--2056. PMLR, 2020.

\bibitem[\protect\citeauthoryear{Cobbe \bgroup \em et al.\egroup
  }{2021}]{cobbe2021phasic}
Karl~W. Cobbe, Jacob Hilton, Oleg Klimov, and John Schulman.
\newblock Phasic policy gradient.
\newblock In {\em International Conference on Machine Learning}, pages
  2020--2027. PMLR, 2021.

\bibitem[\protect\citeauthoryear{Espeholt \bgroup \em et al.\egroup
  }{2018}]{espeholt2018impala}
Lasse Espeholt, Hubert Soyer, Remi Munos, Karen Simonyan, Vlad Mnih, Tom Ward,
  Yotam Doron, Vlad Firoiu, Tim Harley, Iain Dunning, et~al.
\newblock Impala: Scalable distributed deep-rl with importance weighted
  actor-learner architectures.
\newblock In {\em International Conference on Machine Learning}, pages
  1407--1416. PMLR, 2018.

\bibitem[\protect\citeauthoryear{Goyal \bgroup \em et al.\egroup
  }{2017}]{alias2017zforing}
Anirudh Goyal, Alessandro Sordoni, Marc-Alexandre C{\^o}t{\'e}, Nan~Rosemary
  Ke, and Yoshua Bengio.
\newblock Z-forcing: Training stochastic recurrent networks.
\newblock In {\em Advances in neural information processing systems},
  volume~30, 2017.

\bibitem[\protect\citeauthoryear{Gulrajani \bgroup \em et al.\egroup
  }{2017}]{gulrajani2017pixelvae}
Ishaan Gulrajani, Kundan Kumar, Faruk Ahmed, Adrien~Ali Taiga, Francesco Visin,
  David Vazquez, and Aaron Courville.
\newblock Pixel{VAE}: A latent variable model for natural images.
\newblock In {\em International Conference on Learning Representations}, 2017.

\bibitem[\protect\citeauthoryear{Hafner \bgroup \em et al.\egroup
  }{2020}]{hafner2019dream}
Danijar Hafner, Timothy Lillicrap, Jimmy Ba, and Mohammad Norouzi.
\newblock Dream to control: Learning behaviors by latent imagination.
\newblock In {\em International Conference on Learning Representations}, 2020.

\bibitem[\protect\citeauthoryear{Hochreiter and
  Schmidhuber}{1997}]{hochreiter1997long}
Sepp Hochreiter and J{\"u}rgen Schmidhuber.
\newblock Long short-term memory.
\newblock {\em Neural computation}, 9(8):1735--1780, 1997.

\bibitem[\protect\citeauthoryear{Huang \bgroup \em et al.\egroup
  }{2022}]{ppo_implementation}
Shengyi Huang, Rousslan Fernand~Julien Dossa, Antonin Raffin, Anssi Kanervisto,
  and Weixun Wang.
\newblock The 37 implementation details of proximal policy optimization.
\newblock
  \url{https://iclr-blog-track.github.io/2022/03/25/ppo-implementation-details},
  2022.

\bibitem[\protect\citeauthoryear{Ioffe and Szegedy}{2015}]{ioffe2015batch}
Sergey Ioffe and Christian Szegedy.
\newblock Batch normalization: Accelerating deep network training by reducing
  internal covariate shift.
\newblock In {\em International Conference on Machine Learning}, pages
  448--456. PMLR, 2015.

\bibitem[\protect\citeauthoryear{Laskin \bgroup \em et al.\egroup
  }{2020}]{laskin2020curl}
Michael Laskin, Aravind Srinivas, and Pieter Abbeel.
\newblock Curl: Contrastive unsupervised representations for reinforcement
  learning.
\newblock In {\em International Conference on Machine Learning}, pages
  5639--5650. PMLR, 2020.

\bibitem[\protect\citeauthoryear{Laurent \bgroup \em et al.\egroup
  }{2016}]{laurent2016batch}
C{\'e}sar Laurent, Gabriel Pereyra, Phil{\'e}mon Brakel, Ying Zhang, and Yoshua
  Bengio.
\newblock Batch normalized recurrent neural networks.
\newblock In {\em 2016 IEEE International Conference on Acoustics, Speech and
  Signal Processing (ICASSP)}, pages 2657--2661. IEEE, 2016.

\bibitem[\protect\citeauthoryear{LeCun \bgroup \em et al.\egroup
  }{1998}]{lecun1998gradient}
Yann LeCun, L{\'e}on Bottou, Yoshua Bengio, and Patrick Haffner.
\newblock Gradient-based learning applied to document recognition.
\newblock {\em Proceedings of the IEEE}, 86(11):2278--2324, 1998.

\bibitem[\protect\citeauthoryear{Mnih \bgroup \em et al.\egroup
  }{2015}]{mnih2015human}
Volodymyr Mnih, Koray Kavukcuoglu, David Silver, Andrei~A. Rusu, Joel Veness,
  Marc~G. Bellemare, Alex Graves, Martin Riedmiller, Andreas~K. Fidjeland,
  Georg Ostrovski, et~al.
\newblock Human-level control through deep reinforcement learning.
\newblock {\em Nature}, 518(7540):529--533, 2015.

\bibitem[\protect\citeauthoryear{Murphy}{2022}]{pml1Book}
Kevin~P. Murphy.
\newblock {\em Probabilistic Machine Learning: An introduction}.
\newblock MIT Press, 2022.

\bibitem[\protect\citeauthoryear{Nair and Hinton}{2010}]{nair2010rectified}
Vinod Nair and Geoffrey~E. Hinton.
\newblock Rectified linear units improve restricted boltzmann machines.
\newblock In {\em International Conference on Machine Learning}, pages
  807--814, 2010.

\bibitem[\protect\citeauthoryear{Ostrovski \bgroup \em et al.\egroup
  }{2017}]{ostrovski2017count}
Georg Ostrovski, Marc~G Bellemare, A{\"a}ron Oord, and R{\'e}mi Munos.
\newblock Count-based exploration with neural density models.
\newblock In {\em International conference on machine learning}, pages
  2721--2730. PMLR, 2017.

\bibitem[\protect\citeauthoryear{Pathak \bgroup \em et al.\egroup
  }{2017}]{pathak2017icm}
Deepak Pathak, Pulkit Agrawal, Alexei~A. Efros, and Trevor Darrell.
\newblock Curiosity-driven exploration by self-supervised prediction.
\newblock In {\em International Conference on Machine Learning}, pages
  2778--2787. PMLR, 2017.

\bibitem[\protect\citeauthoryear{Pritzel \bgroup \em et al.\egroup
  }{2017}]{pritzel2017neural}
Alexander Pritzel, Benigno Uria, Sriram Srinivasan, Adria~Puigdomenech Badia,
  Oriol Vinyals, Demis Hassabis, Daan Wierstra, and Charles Blundell.
\newblock Neural episodic control.
\newblock In {\em International Conference on Machine Learning}, pages
  2827--2836. PMLR, 2017.

\bibitem[\protect\citeauthoryear{Raffin \bgroup \em et al.\egroup
  }{2021}]{stable-baselines3}
Antonin Raffin, Ashley Hill, Adam Gleave, Anssi Kanervisto, Maximilian
  Ernestus, and Noah Dormann.
\newblock Stable-baselines3: Reliable reinforcement learning implementations.
\newblock {\em Journal of Machine Learning Research}, 22(268):1--8, 2021.

\bibitem[\protect\citeauthoryear{Raileanu and
  Fergus}{2021}]{raileanu2021decoupling}
Roberta Raileanu and Rob Fergus.
\newblock Decoupling value and policy for generalization in reinforcement
  learning.
\newblock In {\em International Conference on Machine Learning}, pages
  8787--8798. PMLR, 2021.

\bibitem[\protect\citeauthoryear{Rosenblatt}{1958}]{rosenblatt1958perceptron}
Frank Rosenblatt.
\newblock The perceptron: a probabilistic model for information storage and
  organization in the brain.
\newblock {\em Psychological review}, 65(6):386, 1958.

\bibitem[\protect\citeauthoryear{Rumelhart \bgroup \em et al.\egroup
  }{1986}]{rumelhart1986learning}
David~E. Rumelhart, Geoffrey~E. Hinton, and Ronald~J. Williams.
\newblock {\em Learning Internal Representations by Error Propagation}.
\newblock MIT Press, 1986.

\bibitem[\protect\citeauthoryear{Schulman \bgroup \em et al.\egroup
  }{2017}]{schulman17ppo}
John Schulman, Filip Wolski, Prafulla Dhariwal, Alec Radford, and Oleg Klimov.
\newblock Proximal policy optimization algorithms.
\newblock {\em arXiv preprint arXiv:1707.06347}, 2017.

\bibitem[\protect\citeauthoryear{Scott}{2010}]{scott2010modern}
Steven~L. Scott.
\newblock A modern bayesian look at the multi-armed bandit.
\newblock {\em Applied Stochastic Models in Business and Industry},
  26(6):639--658, 2010.

\bibitem[\protect\citeauthoryear{Sutton and Barto}{2018}]{SuttonBarto2018}
Richard~S. Sutton and Andrew~G. Barto.
\newblock {\em Introduction to Reinforcement Learning}.
\newblock MIT Press, 2nd edition, 2018.

\bibitem[\protect\citeauthoryear{Tang \bgroup \em et al.\egroup
  }{2017}]{tang2017study}
Haoran Tang, Rein Houthooft, Davis Foote, Adam Stooke, Xi~Chen, Yan Duan, John
  Schulman, Filip De~Turck, and Pieter Abbeel.
\newblock \#{E}xploration: A study of count-based exploration for deep
  reinforcement learning.
\newblock In {\em Advances in Neural Information Processing Systems}, pages
  4--9, 2017.

\bibitem[\protect\citeauthoryear{Tsybakov}{2008}]{tsybakov2008nonparametric}
Alexandre~B. Tsybakov.
\newblock Introduction to nonparametric estimation.
\newblock In {\em Springer Series in Statistics}, 2008.

\bibitem[\protect\citeauthoryear{Zhang \bgroup \em et al.\egroup
  }{2021}]{zhang2021noveld}
Tianjun Zhang, Huazhe Xu, Xiaolong Wang, Yi~Wu, Kurt Keutzer, Joseph~E.
  Gonzalez, and Yuandong Tian.
\newblock Novel{D}: A simple yet effective exploration criterion.
\newblock In {\em Advances in Neural Information Processing Systems}, 2021.

\end{thebibliography}

\appendix
\clearpage

\onecolumn

\setcounter{table}{0}
\renewcommand{\thetable}{A\arabic{table}}
\setcounter{figure}{0}
\renewcommand{\thefigure}{A\arabic{figure}}
\setcounter{algorithm}{0}
\renewcommand{\thealgorithm}{A\arabic{algorithm}}
\setcounter{equation}{0}
\renewcommand{\theequation}{A\arabic{equation}}

\section{Technical Appendix}

Continued from the main text of \textit{DEIR: Efficient and Robust Exploration through Discriminative-Model-Based Episodic Intrinsic Rewards}, 
the technical appendix consist of the following:

\begin{itemize}
    \item \textbf{A.1 Intrinsic Reward Derivation}, which is referred in Section~\ref{sec:episodic-intrinsic-reward}.
    \item \textbf{A.2 Algorithms Used in Modeling}, which is referred in Section~\ref{sec:model-definition}.
    \item \textbf{A.3 Benchmark Environments}, \textbf{A.4 Hyperparameters}, \textbf{A.5 Network Structures}, \textbf{A.6 Computing Environments}, which are referred in Section~\ref{sec:experimental-setup}.
    \item \textbf{A.7 Detailed Experimental Results in MiniGrid}, which is referred in Section~\ref{sec:evaluation-experiments}.
\end{itemize}

\vspace{1em}
\subsection{Intrinsic Reward Derivation}
\label{sec:intrinsic-reward-derivation}

Given a pair of states $(s_t, s_i), \forall i\in[0, t)$, a pair of observations $(o_{t+1}, o_i)$, and an action $a_t$ from the history of an episode, we wish to optimize the agent's policy with an intrinsic reward
\begin{equation}
J=\mathrm{dist}(o_{t+1}, o_i) \cdot I \left( \mathrm{dist}(o_{t+1}, o_i) ; a_t | s_t, s_i \right).
\label{EqOriginalTarget}
\end{equation}

The first term $\mathrm{dist}(o_{t+1}, o_i)$ is the distance between observations $o_{t+1}$ and $o_i$ as used in previous studies~\cite{blundell2016model,pritzel2017neural,badia2020ngu}, and the second term 
$I \left( \mathrm{dist}(o_{t+1}, o_i) ; a_t | s_t, s_i \right)$
is the conditional mutual information between $\mathrm{dist}(o_{t+1}, o_i)$ and $a_t$ given $s_t, s_i$.
Our intuition behind such design is that the policy achieves novelty by maximizing distance between observations $(o_{t+1}, o_i)$ \emph{and} letting the distances to be closely related to $a_t$, the action taken in $s_t$.
This would efficiently eliminate the novelties rooted in the environment's stochasticity other than those brought by the agent's explorations.

\vspace*{0.3em}
Remind that $I(X;Y|Z) = \mathbb{E}_{p(y,z)}\left[\mathrm{D_{KL}}(P(X|Y=y,Z=z)\| P(X|Z=z))\right] $. 
It is because
\begin{align*}
I(X;Y|Z) 
=& \mathbb{E}_{p(x,y,z)}\left[\log \frac{P(X,Y|Z)}{P(X|Z)P(Y|Z)}\right] \\
=& \mathbb{E}_{p(x,y,z)}\left[\log \frac{P(X|Y,Z)\cancel{P(Y|Z)}}{P(X|Z)\cancel{P(Y|Z)}}\right] \\
=& \mathbb{E}_{p(y,z)}\left[\mathbb{E}_{p(x|y,z)} \left[\log \frac{P(X|Y,Z)}{P(X|Z)}\right] \right] \\
=& \mathbb{E}_{p(y,z)}\left[\mathrm{D_{KL}}(P(X|Y=y,Z=z)\| P(X|Z=z))\right]. 
\end{align*}

The second term of $J$ can be expanded as (denoting $D_{t+1,i}\triangleq \mathrm{dist}(o_{t+1}, o_i)$ for simplicity of formulation)
\begin{align*}
I(&D_{t+1,i}; a_t | s_t, s_i) \\
&= \mathbb{E}_{s_t, s_i, a_t, D_{t+1,i}} \left[\mathrm{D_{KL}} (p(D_{t+1,i}, a_t | s_t, s_i) \| p(D_{t+1,i} | s_t, s_i)p(a_t | s_t, s_i) ) \right] \\
% &= \mathbb{E}_{s_t, s_i}\left[\int_{a_t} \int_{D_{t+1,i}} p(D_{t+1,i}, a_t | s_t, s_i) \mathrm{log} \frac{p(D_{t+1,i}, a_t | s_t, s_i)}{p(D_{t+1,i} | s_t, s_i)p(a_t | s_t, s_i)} \diff D_{t+1,i} \diff a_t \right] \\
% &= \mathbb{E}_{s_t, s_i}\left[\int_{a_t} \int_{D_{t+1,i}} p(D_{t+1,i}, a_t | s_t, s_i) \mathrm{log} \frac{p(D_{t+1,i} | s_t, s_i, a_t)p(a_t | s_t, s_i)}{p(D_{t+1,i} | s_t, s_i)p(a_t | s_t, s_i)} \diff D_{t+1,i} \diff a_t \right] \\
% &= \mathbb{E}_{s_t, s_i}\left[\int_{a_t} \int_{D_{t+1,i}} p(D_{t+1,i}, a_t | s_t, s_i) \mathrm{log} \frac{p(D_{t+1,i} | s_t, s_i, a_t)}{p(D_{t+1,i} | s_t, s_i)} \diff D_{t+1,i} \diff a_t \right] \\
% &= \mathbb{E}_{s_t, s_i}\left[\int_{a_t} p(a_t | s_t, s_i) \int_{D_{t+1,i}} p(D_{t+1,i} | s_t, s_i, a_t) \mathrm{log} \frac{p(D_{t+1,i} | s_t, s_i, a_t)}{p(D_{t+1,i} | s_t, s_i)} \diff D_{t+1,i} \diff a_t \right] \\
% &= \mathbb{E}_{s_t, s_i}\left[\int_{a_t} p(a_t | s_t, s_i) \mathrm{D_{KL}}(p(D_{t+1,i} | s_t, s_i, a_t) \| p(D_{t+1,i} | s_t, s_i)) \diff a_t \right] \\
% &= \int_{s_t} \int_{s_i} \int_{a_t} p(s_t, s_i, a_t) \mathrm{D_{KL}}\left(p(D_{t+1,i} | s_t, s_i, a_t) \| p(D_{t+1,i} | s_t, s_i)\right)  \diff a_t \diff s_i \diff s_t \\
&= \mathbb{E}_{s_t, s_i, a_t}\left[ \mathrm{D_{KL}}(p(D_{t+1,i} | s_t, s_i, a_t) \| p(D_{t+1,i} | s_t, s_i)) \right].
\end{align*}

We seek to simplify $J$ which contains an intractable evidence term in the Kullback–Leibler (KL) divergence $D_{KL}$. 
To do so, we devise a surrogate function for the KL divergence by connecting it with total variation (TV).
Denoting $P(x) = p(x | s_t, s_i, a_t)$ and $Q(x) = p(x | s_t, s_i)$, the Bretagnolle–Huber inequality~\cite{bretagnolle1978estimation} gives 
$$
\begin{aligned}
\mathrm{d_{TV}}(P, Q) & \leq \sqrt{1 -\mathrm{exp}(-\mathrm{D_{KL}}(P \| Q))} \\
\Leftrightarrow \; \mathrm{D_{KL}}(P \| Q) & \geq - \mathrm{log} (1 - \mathrm{d^2_{TV}}(P, Q) ),
\end{aligned}
$$
% \\[0.05em]
where $\mathrm{d_{TV}}(P, Q)=\frac{1}{2}\|P - Q\|_1 $ is the total variation between $P$ and $Q$. 
Since 
$\mathrm{d_{TV}}(P, Q)$ 
monotonically increases w.r.t.\ to KL divergence $D_{KL}(P \| Q)$, we use $\mathrm{d_{TV}}$ as the surrogate function.
Furthermore, given $s_t, s_i, a_t$ in deterministic environments (including POMDPs), $s_{t+1}$, $o_{t+1}$ and $D_{t+1,i}$ are uniquely determined.
$P$ is thus a unit impulse function which has the only non-zero value at $D_{t+1,i}$: 
$$
P(x) = p(x | s_t, s_i, a_t) = \begin{cases}
+\infty \; & x = D_{t+1,i}\\
0 \; & x \neq D_{t+1,i}
\end{cases}
\;\;\text{and}\;
\int_{\mathcal{D}}{P(x)dx} = 1,
$$
where $\mathcal{D}$ denotes the set of all possible distances between two observations. 
Also, by the definition of $Q$, we have $\int_{\mathcal{D}}{Q(x)dx} = 1$ and $Q(D_{t+1,i}) \leq 1$.
Given those properties and Scheff{\'e}'s theorem~\cite{tsybakov2008nonparametric}, we obtain 
\begin{align*}
\mathrm{d_{TV}} = 1 - \int_{\mathcal{D}}{\min\bigl(P(x), Q(x)\bigr)dx} = 1 - Q(D_{t+1,i}) .
\end{align*}

We derive the lower bound of the KL divergence using $\mathrm{d_{TV}}$, as
$$
\begin{aligned}
\mathrm{D_{KL}}(P \| Q) 
&\geq - \mathrm{log} (1 - \mathrm{d^2_{TV}}(P, Q))\\
&= - \mathrm{log} (1 - (1 - Q(D_{t+1,i}))^2 )\\
&= - \mathrm{log} (2 \times Q(D_{t+1,i}) - Q(D_{t+1,i})^2 )\\
&\geq - \mathrm{log} (2 \times Q(D_{t+1,i}) )\\
&= - \mathrm{log}{(2)} - \mathrm{log} ( Q(D_{t+1,i}) ) \\
\end{aligned}
$$
where $- \mathrm{log}{(2)}$ is constant, and we effectively maximize the last term $- \mathrm{log} \left( Q(D_{t+1,i}) \right)$.

\begin{figure}[!hbt]
\centering
\includegraphics[height=10.5em]{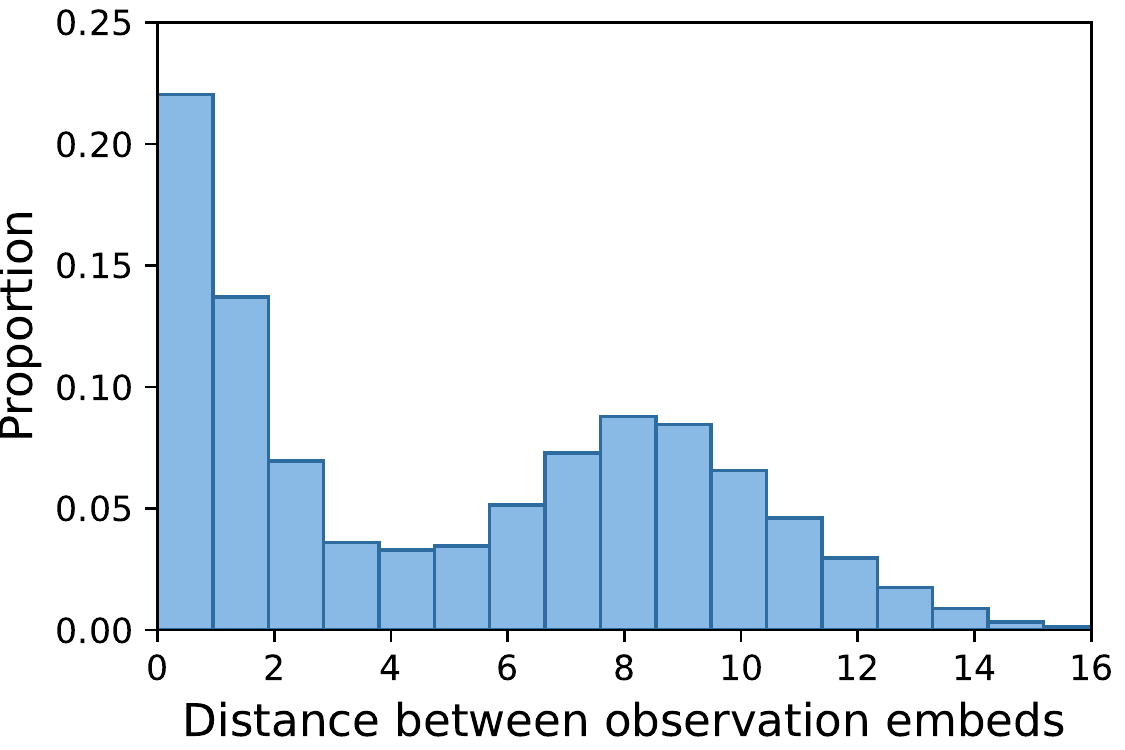}
\caption{
Marginal distribution of $\mathrm{dist}(o_{t+1}, o_i)$ for all pairs of $(o_{t+1}, o_i)$ sampled in a \textit{FourRooms} task with advanced settings, from the start to the middle of training. The first peak (from the left) represents random behaviors at the beginning, and the second reflects directed exploration by partially trained agents. This example supports our assumption that the distance between the current and past observations follows an exponential distribution, especially in early training phases and partially observable environments.
}
\label{figObsDistDistribution}
\end{figure}

Since $D_{t+1,i}\triangleq \mathrm{dist}(o_{t+1}, o_i)$ is the non-negative distance between two observations and $\mathrm{dist}(o_{t+1}, o_i)$ is related to $\mathrm{dist}(s_t, s_i)$, we assume $D_{t+1,i}$ approximately follows an exponential distribution with a mean of the distance between its closest underlying states $s_{t}$ and $s_i$. The intuition here is that limited observability leads to similar observations with minor variations, especially during early training phases when exploration is crucial. The marginal distributions of observation distances observed in our experiments also support this (see Figure~\ref{figObsDistDistribution}).
By setting $D_{t+1,i}|s_t,s_i \sim \mathrm{Exp}({{\lambda}= {1} / {\mathrm{dist}(s_t, s_i)} })$ with an expected value of ${1} / {\lambda} = \mathrm{dist}(s_t, s_i)$, we obtain
\begin{align*}
- \mathrm{log} \bigl( Q(D_{t+1,i}) \bigr) 
&= - \mathrm{log} \left( \lambda \exp{(-\lambda D_{t+1,i})} \right) \\
&= - \mathrm{log} \Bigl( \frac{1}{\mathrm{dist}(s_t, s_i)} \exp{\bigl(-  \frac{\mathrm{dist}(o_{t+1}, o_i)}{\mathrm{dist}(s_t, s_i)} \bigr)} \Bigr) \\
&= \mathrm{log}{\left( \mathrm{dist}(s_t, s_i) \right)} + \frac{\mathrm{dist}(o_{t+1}, o_i)}{\mathrm{dist}(s_t, s_i)}.
\end{align*}

With all of the above, we finally define the lower bound of the objective $J$ as
\begin{equation}
J \ge \mathrm{dist}(o_{t+1}, o_i) \cdot  \mathbb{E}_{s_t, s_i, a_i} \left( \mathrm{log}{\left( \mathrm{dist}(s_t, s_i) \right)} + \frac{\mathrm{dist}(o_{t+1}, o_i)}{\mathrm{dist}(s_t, s_i)} \right).
\label{EqSurrogateTarget}
\end{equation}

Note that the full distributions of observations and states are difficult to sample efficiently. By keeping only the dominating term ${\mathrm{dist}(o_{t+1}, o_i)} / {\mathrm{dist}(s_t, s_i)}$ and relaxing the expectation to the minimum, we further simplify Equation~\ref{EqSurrogateTarget} to
\begin{equation}
J \ge \min_{i}{\frac{\mathrm{dist}^2(o_{t+1}, o_i)}{\mathrm{dist}(s_t, s_i)}},
\label{EqSurrogateTargetSimplified}
\end{equation}
where $o_{t+1}, o_i$, and $s_t, s_i$ are random variables in Equations~\ref{EqOriginalTarget} and \ref{EqSurrogateTarget} but realized values in Equation~\ref{EqSurrogateTargetSimplified}.
We find that Equation~\ref{EqSurrogateTargetSimplified} is much simpler than Equation~\ref{EqSurrogateTarget}, yet delivers a performance that is just as good or even better.
Equation~\ref{EqSurrogateTargetSimplified} is equivalent to the intrinsic reward proposed in the main text (Equation~\ref{EqProposedIntrinsicReward}).

\vspace*{0.5em}

\newpage

\subsection{Algorithms Used in Modeling}
\label{sec:algorithms-used-in-modeling}

In Algorithm~\ref{alg:algoIRGen}, we present the pseudocode for generating the proposed episodic intrinsic reward $r^I_t$ at time step $t$ and maintaining the episodic history of observation and trajectory embeddings .
In Algorithm~\ref{alg:algoObsQueue}, the proposed DEIR maintains a queue of recent observations in memory, storing novel observations collected in the most recent episodes. 
The pseudocode about how DEIR updates the observation queue $\mathbb{Q}$, and which kinds of observations are considered ``novel'' during the updating.

\begin{algorithm}[!htb]
\caption{Generating intrinsic reward $r^I_t$ at time step $t$}
\label{alg:algoIRGen}
\textbf{Input}: 
$\eobs_t$, $E_{\textsc{\tiny obs}}=\{\eobs_0, \cdots, \eobs_{t-1}\}$,  \\
\hspace*{2.7em} $\etraj_t$, $E_{\textsc{\tiny traj}}=\{\etraj_0, \cdots, \etraj_{t-1}\}$,  \\
\hspace*{2.7em} $\textit{terminal}(s_t)$  \\
%\textbf{Parameter}: Optional list of parameters\\
\textbf{Output}: $r^I_t$ 
\begin{algorithmic}[1] %[1] enables line numbers
\IF {$E_{\textsc{\tiny obs}} = \emptyset$}
\STATE \textbf{return} $0$
\ENDIF

\STATE $d_\textsc{\tiny obs} \leftarrow$ Euclidean distances between the episodic observation embedding history $E_{\textsc{\tiny obs}}$ and $\eobs_t$.
\STATE $d^2_\textsc{\tiny obs} \leftarrow$ element-wise squared $d_\textsc{\tiny obs}$.
\STATE $d_\textsc{\tiny traj} \leftarrow$ Euclidean distances between the episodic trajectory embedding history $E_{\textsc{\tiny traj}}$ and $\etraj_t$.
\STATE $d_\textsc{\tiny joint} \leftarrow d^2_\textsc{\tiny obs} / d_\textsc{\tiny traj}$ (element-wise).

\IF {$\textit{terminal}(s_t)$}
\STATE $E_{\textsc{\tiny obs}} \leftarrow \emptyset$
\STATE $E_{\textsc{\tiny traj}} \leftarrow \emptyset$
\ELSE
\STATE $E_{\textsc{\tiny obs}} \leftarrow E_{\textsc{\tiny obs}} \cup \eobs_t$
\STATE $E_{\textsc{\tiny traj}} \leftarrow E_{\textsc{\tiny traj}} \cup \etraj_t$
\ENDIF
\STATE \textbf{return} minimum element in $d_\textsc{\tiny joint}$
\end{algorithmic}
\end{algorithm}

\begin{algorithm}[!htb]
\caption{Updating the queue of recent observations $\mathbb{Q}$ (from which to randomly select fake training data)}
\label{alg:algoObsQueue}
\textbf{Input}: 
$\mathbb{Q}$, $o_t$, $r^I_t$, running average of $r^I$ \\
\textbf{Parameter}: maximum size of $\mathbb{Q}$, smoothing factor for updating the running average of intrinsic rewards\\
\textbf{Output}: $\mathbb{Q}'$
\begin{algorithmic}[1] %[1] enables line numbers

\STATE Update the running average of $r^I$ with $r^I_t$.
\IF {$\mathbb{Q}=\emptyset$ or $r^I_t \geq$ the running average of $r^I$}
\IF {$|\mathbb{Q}| = $ maximum size of $\mathbb{Q}$}
\STATE $\mathbb{Q}' \leftarrow \mathbb{Q}$ with the oldest element being removed
\STATE $\mathbb{Q}' \leftarrow \mathbb{Q}'\cup o_t$
\ELSE
\STATE $\mathbb{Q}' \leftarrow \mathbb{Q}\cup o_t$
\ENDIF
\ENDIF

\STATE \textbf{return} $\mathbb{Q}'$
\end{algorithmic}
\end{algorithm}

\clearpage

In Algorithm~\ref{alg:algo3x3}, Algorithm~\ref{alg:algoNoisyObs} and Algorithm~\ref{alg:algoNoWall} we describe three types of modifications applied to the environment's observations in advanced MiniGrids games. See Section~\ref{sec:robustness-exp} for descriptions about all the three modifications.

\begin{algorithm}[!ht]
\caption{Processing observations for MiniGrid variants of reduced view sizes}
\label{alg:algo3x3}
\textbf{Input}: original $o_t$ ($7 \times 7$ view size)\\
\textbf{Output}: processed $o'_t$ ($3 \times 3$ view size)\\[-1.2em]
\begin{algorithmic}[1] %[1] enables line numbers
\STATE Let $(x_{\textit{\tiny agent}}, y_{\textit{\tiny agent}})$ be the fixed coordinates of the agent
\STATE $o'_t \leftarrow$ $o_t$ being cropped into a $3 \times 3$ image centered on $(x_{\textit{\tiny agent}}, y_{\textit{\tiny agent}} + 1)$ (to always show the agent on the center bottom)
\STATE \textbf{return} $o'_t$
\end{algorithmic}
\end{algorithm}

\begin{algorithm}[!ht]
\caption{Processing observations for MiniGrid variants of noisy observations}
\label{alg:algoNoisyObs}
\textbf{Input}: original $o_t$, $\mu$, $\sigma$\\
\textbf{Parameter}: the mean and standard deviation of noise\\
\textbf{Output}: processed $o'_t$\\[-1.2em]
\begin{algorithmic}[1] %[1] enables line numbers
\STATE $ o_{\textit{\tiny noise}} \leftarrow $ Randomly generated image with $7 \times 7 \times 3$ normally distributed noise (mean: $\mu$, standard deviation: $\sigma$)
\STATE $o'_t \leftarrow o_t +  o_{\textit{\tiny noise}}$
\STATE \textbf{return} $o'_t$
\end{algorithmic}
\end{algorithm}

\begin{algorithm}[!ht]
\caption{Processing observations for MiniGrid variants of invisible obstacles}
\label{alg:algoNoWall}
\textbf{Input}: original $o_t$\\
\textbf{Output}: processed $o'_t$\\[-1.2em]
\begin{algorithmic}[1] %[1] enables line numbers
\STATE $o'_t \leftarrow$ $o_t$ with all \textit{wall} objects being removed
\STATE $o'_t \leftarrow$ $o'_t$ with the colors of \textit{unseen}, \textit{empty} and \textit{wall} grids being changed to the color of the \textit{floor} grid
\STATE \textbf{return} $o'_t$
\end{algorithmic}
\end{algorithm}

\clearpage

\subsection{Benchmark Environments}
\label{sec:benchmark-environments}

We describe each task environment used in our experiments. More detailed introductions can be found on the official project homepage of each benchmark \cite{gym_minigrid,procgen_github,cobbe2020procgen}.

\subsubsection{Standard MiniGrid Environments}
\begin{itemize}

\item \textit{FourRooms}: Given four $8 \times 8$ rooms separated by walls, each room is connected with other two rooms through a $1 \times 1$ gap in the wall. An agent starts from one room and moves towards the goal in another room. The initial positions of the agent, goal and gaps are randomly selected. This task doesn't include any door or key.

\item \textit{MultiRoom-N2-S4}: The task consists of two adjacent rooms to explore. The rooms are connected by a closed but unlocked door. An agent starts from the first room, moving towards the goal in the last room. The maximum size of each room is $4 \times 4$. The number of all generated grids is $25 \times 25$ in total.

\item \textit{MultiRoom-N4-S5}: The task consists of four cascaded rooms to explore. Any two adjacent rooms are connected by a closed but unlocked door. An agent starts from the first room, moving towards the goal in the last room. The maximum size of each room is $5 \times 5$. The size of all generated grids is $25 \times 25$.

\item \textit{MultiRoom-N6}: The task is similar to \textit{MultiRoom-N4-S5}, but with six cascaded rooms. The maximum size of each room is $10 \times 10$. The size of all generated grids is $25 \times 25$.

\item \textit{MultiRoom-N30}: A customized \textit{MultiRoom} task, with 30 cascaded rooms. The maximum size of each room is $10 \times 10$. The size of all generated grids is $45 \times 45$.

\item \textit{DoorKey-8x8}: Two adjacent rooms are separated by a wall. An agent needs to find a key in the first room, unlock and open a door in the wall, and reach the goal in the second room. The total size of the two rooms is $8 \times 8$. The initial positions of the agent, key, wall, and door are randomly generated. The goal is fixed at the bottom right of the second room.

\item \textit{DoorKey-16x16}: The task is similar to \textit{DoorKey-8x8}, but with a total size of $16 \times 16$.

\item \textit{KeyCorridorS4R3}: The task consists of a corridor and four rooms divided by walls and doors. An agent starting from the corridor in the center needs to find a key from an unlocked room and get to the goal hidden behind another locked door. 

\item \textit{KeyCorridorS6R3}: The task setting is similar to \textit{KeyCorridorS4R3}, but with six rooms to explore in total.

\item \textit{ObstructedMaze-Full}: The task consists of $3 \times 3$ rooms, each of which is connected by two or more doors and may have multiple interactable objects inside. A ball in a random color is placed in one of the four corner rooms. The agent is required to pick up the ball as its target. To do so, it needs to open locked doors using keys hidden in boxes, and move away balls set in front of doors as obstacles when necessary. 

\end{itemize}

\subsubsection{Advanced MiniGrid Environments}
We show the pseudocode for processing observations in the three advanced MiniGrid variants of reduced view sizes, noisy observations and invisible obstacles in Algorithm~\ref{alg:algo3x3}, \ref{alg:algoNoisyObs} and \ref{alg:algoNoWall}, respectively.
Note that the agent's view size can be specified as an option when registering the environment in MiniGrid, so we didn't re-implement the resizing function. The pseudocode shown in \ref{alg:algo3x3} is mainly for a better description.

\subsubsection{ProcGen Environments}

\begin{itemize}

\item \textit{Ninja}: A side-scroller game with multiple narrow ledges randomly placed at different horizontal levels. The agent starting from the leftmost needs to jump across those ledges and get to the goal placed on the top of a ledge on the right. When moving towards the goal, the agent must either avoid bomb obstacles or destroy them by attacks. Taking ``jump'' actions over multiple timesteps can let the agent jump higher. The maximum reward is 10 for reaching the goal.

\item \textit{Climber}: A vertical platform game. The agent starts at the bottom of the screen, and needs to climb a number of platforms to get to the top while collecting as many stars as possible along the way. The agent can get a small reward by collecting a star, and get a larger reward after collecting all stars in an episode. An episode ends after all stars are collected or the agent is caught by flying monsters scattered throughout the game.

\item \textit{Jumper}: An open-world game with an agent, a goal, obstructive terrains, and stationary enemies. All objects in the world are completely randomly placed. The agent may need to ascend or descend its horizontal level and avoid obstacles and enemies by jumping. The agent is also capable of ``double jumping'', namely, it can jump the second time while in the air. A reward of 10 is given after reaching the goal.

\item \textit{Caveflyer}: An open-world game, in which the agent needs to explore multiple connected caves to find an exit (the goal). The agent can rotate and move forward or backward, as in a physical environment with both velocity and acceleration. In addition to obstructive terrains, there might also be moving enemies and bonus targets. The agent can get extra rewards by destroying bonus targets (3 points per target) and a fixed completion reward of 10 for reaching the goal. 

\end{itemize}
Each ProcGen task has an option to specify the difficulty of levels to generate.
We conducted all experiments on the hard mode.
Compared with easy mode games, hard mode games are larger in size and have more obstacles and enemies, and thus are harder for the agent to obtain rewards.

\newpage

\begin{table*}[!p]
\centering

\begin{tabular}{llllp{6cm}}
    \hline \\[-1.1em]
        Hyperparameter & MiniGrid & OMFull $^{(*)}$ & ProcGen & Candidate Values \\ \\[-1.1em]
    \hline  \\[-0.8em]
	PPO $\gamma$ & 0.99 & 0.99 & 0.99 & 0.99, 0.997, 0.999\\ \\[-1.1em]
	PPO $\lambda_{\tiny \textsc{GAE}}$ & 0.95 & 0.95 & 0.95 & 0.0, 0.8, 0.9, 0.95, 0.99, 0.997 \\ \\[-1.1em]
	PPO rollout steps & 512 & 256 & 256 & 128, 256, 512, 1024, 2048, 4096 \\ \\[-1.1em]
	PPO workers & 16 & 64 & 64 & 16, 32, 64, 128 \\ \\[-1.1em]
	PPO clip range & 0.2 & 0.2 & 0.2 & 0.1, 0.2, 0.3 \\ \\[-1.1em]
	PPO training epochs & 4 & 3 & 3 & 2, 3, 4 \\ \\[-1.1em]
	model training epochs & 4 & 3 & 3 & 1, 3, 4 \\ \\[-1.1em]
	mini-batch size & 512 & 2048 & 2048 & 512, 1024, 2048 \\ \\[-1.1em]
	% policy gradient coef & 1.0 & 1.0 & 1.0 & 1.0 \\ \\[-1.1em]
	% value loss coef & 0.5 & 0.5 & 0.5 & 0.5, 1.0 \\ \\[-1.1em]
	entropy loss coef & \num{1e-2} & \num{5e-4} & \num{1e-2} & \num{5e-4}, \num{1e-3}, \num{1e-2}, \num{1e-1} \\ \\[-1.1em]
	advantage normalization & yes & no & no & yes, no \\ \\[-1.1em]
	adv norm momentum & 0.9 & - & - & 0.0, 0.9, 0.95, 0.99, 0.999 \\ \\[-1.1em]
	Adam learning rate & \num{3e-4} & \num{1e-4} & \num{1e-4} & \num{3e-5}, \num{1e-4}, \num{3e-4}, \num{1e-3} \\ \\[-1.1em]
	Adam epsilon & \num{1e-5} & \num{1e-5} & \num{1e-5} & \num{1e-2}, \num{1e-4}, \num{1e-5}, \num{1e-8}\\ \\[-1.1em]
	Adam beta1 & 0.9 & 0.9 & 0.9 & 0.9, 0.95, 0.99 \\ \\[-1.1em]
	Adam beta2 & 0.999 & 0.999 & 0.999 & 0.999, 0.9999 \\ \\[-1.1em]
	normalization for layers & Batch Norm & Layer Norm & Layer Norm & Batch Norm, Layer Norm, no \\ \\[-1.1em]
        seeds in experiments  & 20 & 12 & 3 & \small{(decided per our computing capabilities)}\\ \\[-1.1em]
        extrinsic reward coef  & 1.0 & 10.0 & 1.0 & \\ \\[-0.8em]
	% BN momentum & 0.9 & - & 0.9, 0.99, 0.997 \small{(= 0.1, 0.01, 0.003 in PyTorch)} \\ \\[-1.1em]
	% dropout & no & no & 0.1, 0.2, no \\ \\[-1.1em]
	% regularization & no & no & \num{1e-4}, \num{1e-3}, \num{1e-2}, no \\ \\[-0.8em]
    \hline \\[-0.8em]
	\textit{DEIR} & & & \\ \\[-1.1em]
	IR (intrinsic reward) coef $\beta$ & \num{1e-2} & \num{1e-3} & \num{5e-2} $^{(**)}$ & \num{1e-3}, \num{3e-3}, \num{5e-3}, \num{1e-2}, \num{3e-2}, \num{5e-2}, \num{1e-1} \\ \\[-1.1em]
	IR normalization & yes & yes & yes & yes, no \\ \\[-1.1em]
	IR norm momentum & 0.9 & 0.9 & 0.9 & 0.0, 0.9, 0.95, 0.99, total average \\ \\[-1.1em]
	observation queue size & \num{1e5} & \num{1e5} & \num{1e5} & \num{1e4}, \num{1e5}, \num{1e6} \\ \\[-0.7em]
	\textit{NovelD} & & & \\ \\[-1.1em]
	IR coefficient $\beta$ & \num{3e-2} & \num{3e-3} & \num{3e-2} & \num{1e-4}, \num{1e-3}, \num{3e-3}, \num{5e-3}, \num{1e-2}, \num{3e-2}, \num{5e-2}, \num{1e-1} \\ \\[-1.1em]
	IR normalization & yes & yes & yes & yes \\ \\[-1.1em]
	IR norm momentum & 0.9 & 0.9 & 0.9 & 0.9 \\ \\[-1.1em]
	RND error normalization& no & no & no & yes, no \small{(using IR norm only performed better)} \\ \\[-1.1em]
	RND error momentum & total avg & total avg & total avg & 0.997, total average \\ \\[-0.7em]
	\textit{NGU} & & & \\ \\[-1.1em]
	IR coefficient $\beta$ & \num{1e-3} & - & \num{3e-4} & \num{1e-4}, \num{3e-4}, \num{1e-3}, \num{3e-3}, \num{5e-3}, \num{1e-2}, \num{3e-2}, \num{1e-1} \\ \\[-1.1em]
	IR normalization & yes & - & yes & yes, no \\ \\[-1.1em]
	IR norm momentum & 0.0 & - & 0.0 & 0.0, 0.2, 0.5, 0.9 \\ \\[-1.1em]
	RND-based lifelong bonus & yes & - & yes & yes, no \\ \\[-1.1em]
	RND error normalization & yes & - & yes & yes, no \\ \\[-1.1em]
	RND error momentum & total avg & - & total avg & 0.997, total average \\ \\[-1.1em]
	momentum (squared dist.) & 0.997 & - & 0.997 & 0.99, 0.997, 0.999, 0.9999, total average \\ \\[-0.7em]
	\textit{RND} & & & \\ \\[-1.1em]
	IR coefficient $\beta$ & \num{3e-3} & - & \num{1e-2} & \num{1e-5}, \num{1e-4}, \num{1e-3}, \num{3e-3}, \num{5e-3}, \num{1e-2}, \num{3e-2} \\ \\[-1.1em]
	IR normalization & yes & - & yes & yes, no \\ \\[-1.1em]
	IR norm momentum & 0.9 & - & 0.9 & 0.0, 0.9 \\ \\[-1.1em]
	RND error normalization& no & - & no & yes, no \small{(using IR norm only performed better)} \\ \\[-1.1em]
	RND error momentum & total avg & - & total avg & 0.997, total average \\ \\[-0.7em]
	\textit{ICM} & & & \\ \\[-1.1em]
	IR coefficient $\beta$ & \num{1e-2} & - & \num{1e-4} & \num{3e-5},\num{1e-4}, \num{3e-4}, \num{1e-3}, \num{3e-3}, \num{5e-3}, \num{1e-2}, \num{3e-2} \\ \\[-1.1em]
	IR normalization & yes & - & yes & yes \\ \\[-1.1em]
	IR norm momentum & 0.9 & - & 0.9 & 0.9 \\ \\[-1.1em]
	forward loss coef. & 0.2 & - & 0.2 & 0.2 \small{(following the original paper)} \\ \\[-1.1em]
    \hline
\end{tabular}\\[-0.4em]
\caption{
Hyperparameters used for training each method in MiniGrid and ProcGen. Hyperparameters were searched in \textit{DoorKey-8x8} and \textit{Jumper}, and partially adjusted according to results in \textit{DoorKey-16x16}, \textit{KeyCorridorS6R3}, \textit{ObstructedMaze-Full} and \textit{Caveflyer} tasks. Values performed best across all tests were selected and used in the remaining experiments.
The term ``momentum'' refers to the smoothing factor in the exponential moving average (EMA) used to calculate the running mean and variance.
The listed values contain all values that have been searched and tested in at least one experiment.
Due to limited computing resources, we only search a subset of all possible combinations of the listed parameter values.
(* OMFull refers to \textit{ObstructedMaze-Full} in MiniGrid, ** DEIR uses $\beta=$ \num{5e-3} in \textit{Caveflyer})
}
\label{tableHyperParams}
\end{table*}

\subsection{Hyperparameters}
\label{sec:hyperparameters}

Hyperparameters used for training each method in MiniGrid and ProcGen environments are summarized in Table~\ref{tableHyperParams}.
\subsubsection{Hyperparameter Selection}
In the experiments, all methods are implemented on top of PPO.  Therefore, the hyperparameters of PPO were shared among all methods while the hyperparameters of each method, including intrinsic reward coefficient were tuned separately.  
%%
%Considering that all methods experimented with in this paper are reward-based exploration methods, we tested and selected the same set of hyperparameters for their common base learning algorithm (PPO). 
%%
%Then, based on the common PPO hyperparameters, we conducted experiments to find the best intrinsic reward coefficient and other method-specific parameters that can maximize the performance of each corresponding method, respectively.
%
Hyperparameters were initially searched in \textit{DoorKey-8x8} and \textit{Jumper} tasks for MiniGrid and ProcGen experiments, and then partially adjusted per the results in \textit{DoorKey-16x16}, \textit{KeyCorridorS6R3}, \textit{ObstructedMaze-Full} (\textit{OMFull}) and \textit{Caveflyer} tasks.
The values that consistently performed well across all tests were selected, and used in the experiments.
For methods that failed to learn any effective policy in difficult tasks, we selected hyperparameters for them according to their best records in simpler tasks.

\subsubsection{Normalization for Hidden Layers}
We applied batch normalization \cite{ioffe2015batch} to all non-RNN hidden layers used in MiniGrid experiments (except for \textit{OMFull}), and applied layer normalization \cite{ba2016layer} to all CNN layers in ProcGen experiments (and in \textit{OMFull}).
It is because that in tasks with stable observation distributions, we found that applying batch normalization to all non-RNN layers could significantly boost up the learning speed. 
Similar results were also observed when applying batch normalization to RNN layers. However, we only adopted batch normalization for non-RNN layers in order to avoid the risk of causing serious overfitting~\cite{laurent2016batch}.
In tasks with ever-changing observations like ProcGen games, we applied layer normalization to all CNN layers, together with a smaller learning rate to learn policies in a more gradual way.

\subsubsection{Normalization for Intrinsic Rewards}
Following RND and NGU, we normalized intrinsic rewards for all methods using the formula
$(r^I_t - \mu_{\tiny \textit{IR}}) / \sigma_{\tiny \textit{IR}}$, where $\mu_{\tiny \textit{IR}}$ and $\sigma_{\tiny \textit{IR}}$ are respectively the exponential moving average and standard deviation of intrinsic rewards. 
$\mu_{\tiny \textit{IR}}$ and $\sigma_{\tiny \textit{IR}}$ were updated using all samples collected in one RL roll-out, and the momentum values for old $\mu_{\tiny \textit{IR}}$ and $\sigma_{\tiny \textit{IR}}$ were set as $0.9$ except for NGU. It was because NGU already normalized the squared Euclidean distances when generating intrinsic rewards.

\subsubsection{Normalization for Advantages}
In MiniGrid experiments, we found advantage normalization \cite{ppo_implementation,andrychowicz2021matters} was helpful in improving both learning speed and exploration efficiency, and we applied it in all MiniGrid experiments. 
The momentum for computing exponential moving averages and standard deviations was 0.9.
In ProcGen experiments, we also found that advantage normalization could facilitate early exploration, but the agent's policy was occasionally trapped in local optima when it was enabled.
Since this problem was not observed in MiniGrid, we applied advantage normalization in MiniGrid only.

\subsubsection{Maximum Episode Length}
For the experiments reported in the main paper and in Figures~\ref{figFullMinigridReturn}, \ref{figFullMinigridEpExpl} and \ref{figFullMinigridLlExpl}, we simply used the default maximum episode length set by the benchmark and observed that DEIR outperforms other methods in most tasks.
We additionally evaluated whether DEIR can perform well with other maximum episode lengths, as an episodic exploration method.
We compared its performance in \textit{MultiRoom-N30} environments with maximum episode lengths of 300, 500, 1000, 2000, and 5000. (The default maximum episode length is 20 times of the number of rooms in standard \textit{MultiRoom} games, and thus is 600 in \textit{MultiRoom-N30}.)
Besides the fixed completion reward of $1.0$, the original MiniGrid environment also gives a time penalty to encourage completing the game using fewer steps. 
The time penalty is computed based on the ratio of used steps and maximum steps, and thus may be affected when changing the maximum episode length.
For fair comparisons, we temporarily disabled the time penalty from the environment.

\begin{figure*}[!h]
\centering
\includegraphics[height=13.1em]{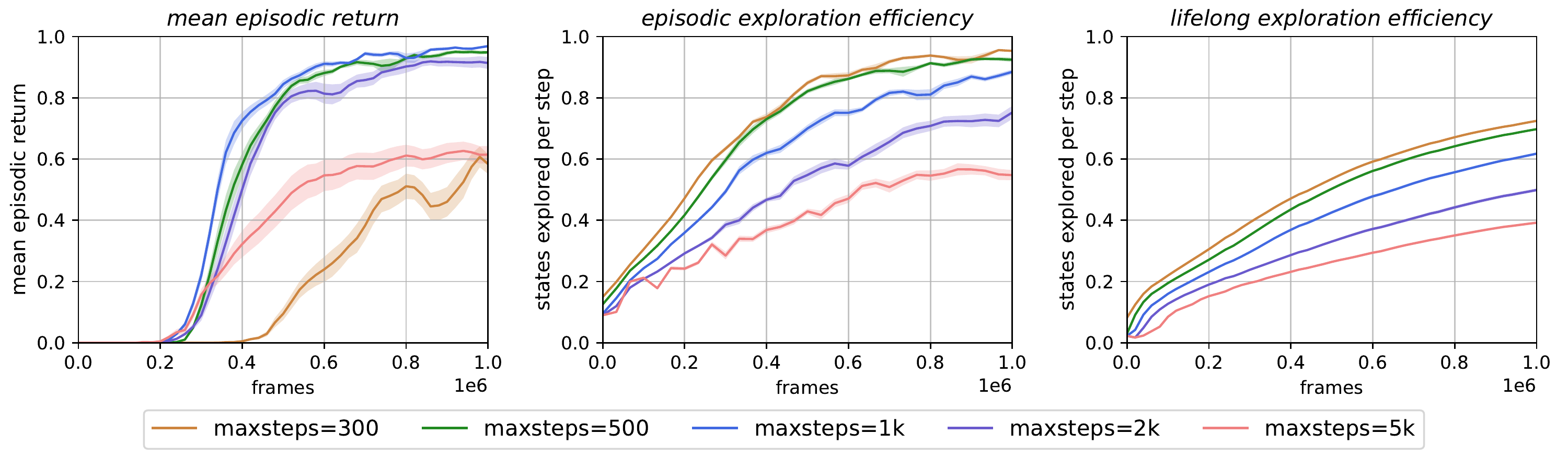}
\caption{
Results on maximum episode length. 
In the \textit{MultiRoom-N30} environment, DEIR performs well with a relatively wide range of maximum episode lengths (e.g., 500, 1000, and 2000). 
Too short lengths (e.g., 300) may cause the agent to be less likely to reach the goal and thus focus only on obtaining intrinsic rewards.
Too long lengths (e.g., 5000) may prevent the agent from collecting sufficient intrinsic rewards in later steps of the episode.} 
\label{figMaxSteps}
\end{figure*}

The results in Figure~\ref{figMaxSteps} demonstrate that DEIR can perform robustly within a wide range of maximum episode lengths, at least ranging from 500 to 2000 and covering the default maximum length of 600. 
Also, we believe DEIR can perform even better if the maximum episode length is specifically optimized or dynamically managed.

\subsubsection{Maximum Observation Queue Size}

Another hyperparameter inclued in our proposal of DEIR is the maximum length of the first-in-first-out queue $\mathbb{Q}$ -- as described in Section~\ref{sec:model-definition} and Algorithm~\ref{alg:algoObsQueue}, $\mathbb{Q}$ stores recent novel observations to make fake input data for the discriminative model's training. 
In order to make clear the most suitable value for it, we conducted experiments on different maximum sizes for $\mathbb{Q}$.  

\begin{figure}[!h]
\centering
\includegraphics[height=12.5em]{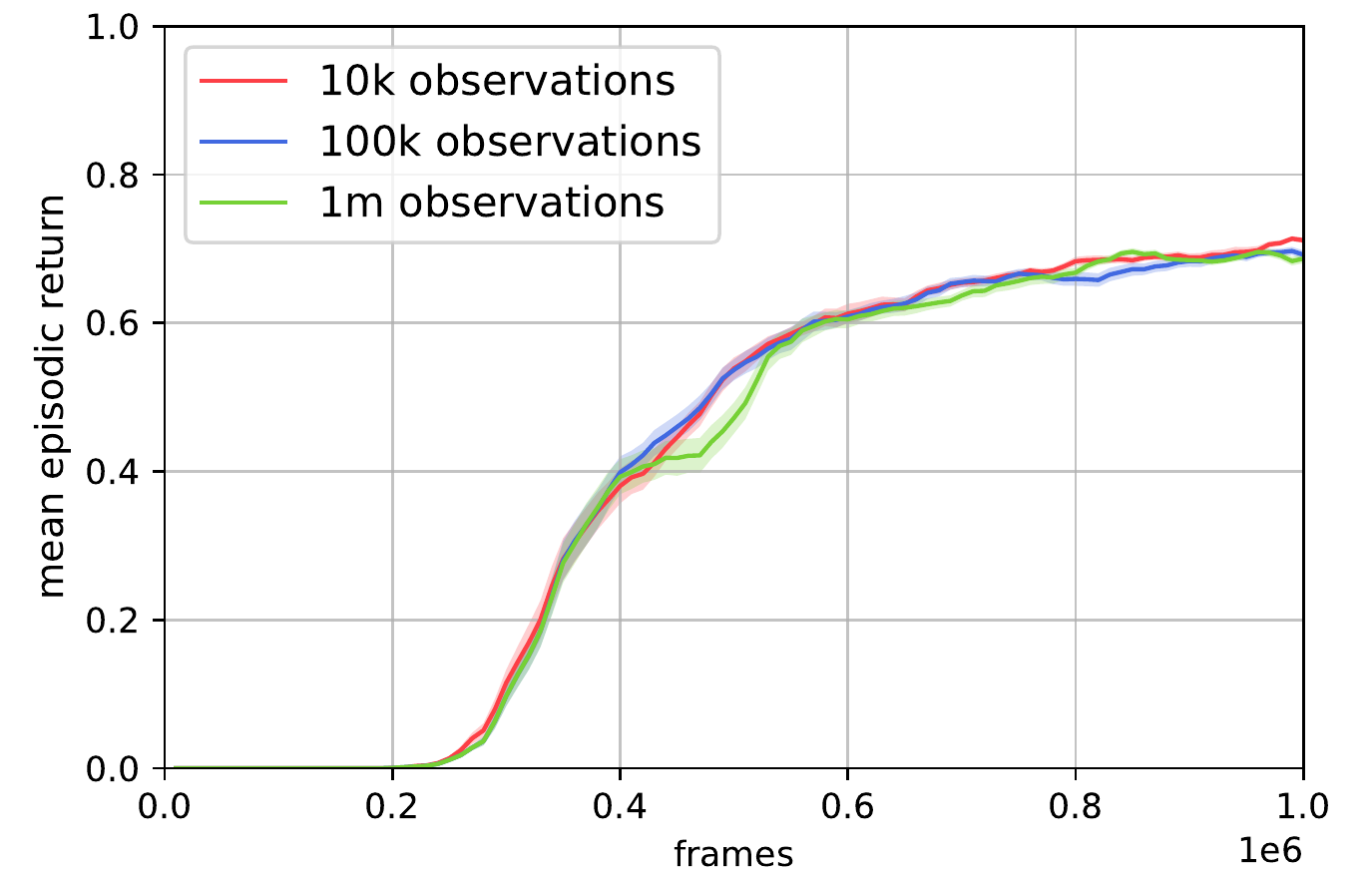}
\caption{
Comparing mean episodic returns when using different maximum sizes for the recent observation queue $\mathbb{Q}$.
The results show that DEIR is compatible with a wide range of observation queue sizes, ranging from \num{1e4} to \num{1e6}.
}
\label{figObsqueue}
\end{figure}

The results shown in Figure~\ref{figObsqueue} suggest that, in the \textit{MultiRoom-N30} environment, DEIR can actually work well with a wide range of observation queue sizes, ranging from \num{1e4} to \num{1e6}. Nonetheless, in order to save computing resources while keeping better diversity, we adopted \num{1e5} in the main experiments.

\newpage

\subsection{Network Structures}
\label{sec:network-structures}

We define the structures of policy and value networks, and the proposed discriminative model, which were updated separately in parallel to the training of PPO agents.
Policy and value networks shared the same CNN and RNN modules, but used separate MLP output heads.
Specifically for CNNs: 
In MiniGrid experiments, we used the CNN structure simplified from the CNN used in DQN~\cite{mnih2015human} for both the policy and the model. 
In ProcGen experiments, we used the same CNN structure as in IMPALA~\cite{espeholt2018impala} and Cobbe et al.'s work~\cite{cobbe2020procgen} for the agent's policy and value function, and a widened version of the CNN used in DQN for the dynamics model.

\subsubsection{Policy and value networks (MiniGrid)} 

\begin{description}

% MiniGrid

\item[CNN (view size $7 \times 7$)]\ \\
Conv2d(in=3, out=32, kernel=2, stride=1, pad=0),  \\
Conv2d(in=32, out=64, kernel=2, stride=1, pad=0),  \\
Conv2d(in=64, out=64, kernel=2, stride=1, pad=0),  \\
FC(in=1024, out=64).

\item[CNN (view size $3 \times 3$)]\ \\
Conv2d(in=3, out=32, kernel=2, stride=1, pad=1), \\
Conv2d(in=32, out=64, kernel=2, stride=1, pad=0), \\
Conv2d(in=64, out=64, kernel=2, stride=1, pad=0), \\
FC(in=256, out=64).

\item[RNN]\ \\
GRU(in=64, out=64).

\item[MLP (value network)]\ \\
FC(in=64, out=128),\\
FC(in=128, out=1).

\item[MLP (policy network)]\ \\
FC(in=64, out=128),\\
FC(in=128, out=number of actions).

\end{description}
\textit{FC} stands for the fully connected linear layer, and \textit{Conv2d} refers to the 2-dimensional convolutional layer, \textit{GRU} is the gated recurrent units \cite{cho2014gru}.
In MiniGrid games, batch normalization and ReLU~\cite{nair2010rectified} are applied after each hidden layer listed above, except for the layers inside the RNN modules. Batch normalization is also used for normalizing input images.

\subsubsection{Discriminative model (MiniGrid)}
\begin{description}
\item[CNN] Same structure as the policy and value networks.
\item[RNN] Same structure as the policy and value networks.
\item[MLP] \ \\
FC(in=64 $\times$ 2 + number of actions, out=128), \\
FC(in=128, out=128), \\
FC(in=128, out=1).
\end{description}
Batch normalization and ReLU are applied after each hidden layer, except for RNN layers. Batch normalization is also applied to input images.

\subsubsection{Policy and value networks (Procegn)} 

\begin{description}

\item[CNN]\ \\
Same structure as the CNN used in IMPALA,\ \\
FC(in=2048, out=256).

\item[RNN]\ \\
GRU(in=256, out=256).

\item[MLP (value network)]\ \\
FC(in=256, out=256), \\
FC(in=256, out=1).

\item[MLP (policy network)]\ \\
FC(in=256, out=256),\\
FC(in=256, out=number of actions).
\end{description}
In ProcGen games (and \textit{ObstructedMaze-Full} in MiniGrid), layer normalization and ReLU are applied after each CNN layer. Layer normalization is also used for input images.

\subsubsection{Discriminative model (ProcGen)}
\begin{description}
\item[CNN]\ \\
Conv2d(in=3, out=32, kernel=8, stride=4, pad=0), \\
Conv2d(in=32, out=64, kernel=4, stride=2, pad=0), \\
Conv2d(in=64, out=64, kernel=4, stride=1, pad=0), \\
FC(in=576, out=256).

\item[RNN]\ \\
GRU(in=256, out=256).

\item[MLP] \ \\
FC(in=256 $\times$ 2 + number of actions, out=256), \\
FC(in=256, out=256), \\
FC(in=256, out=1).
\end{description}
Layer normalization and ReLU are applied after each CNN layer. Layer normalization is also used for input images.

\subsubsection{Other Implementation Details}
For fair comparisons, we basically adopted the same model structure as the discriminative model for the networks of existing exploration methods used in our experiments, except for RND and RND-based methods. 
In RND, NovelD and NGU, the MLP heads of the target and predictor networks were larger than those in other methods, in order to alleviate the problem of vanishing lifelong rewards.

For ICM, RND, and NovelD, we reproduced those methods based on their original papers and open-source code, but adopted newly-tuned hyperparameters and network structures when applicable.
For NGU, the NGU method in our experiments is not entirely identical to the original implementation. It is because the learning framework we adopted for training and comparing all exploration methods is more lightweight than the originally proposed one. However, we confirmed that every step of the original algorithm for intrinsic reward generation was followed in our reproduction (according to the original definition given in NGU's paper). 

Our PPO agents were created based on the implementation of PPO in Stable Baselines 3 (1.1.0 version)~\cite{stable-baselines3}.
For more details on our implementations of DEIR and existing methods, please refer to our source code which is available at \url{https://github.com/swan-utokyo/deir}.

\newpage
\subsection{Computing Environments}
\label{sec:computing-environments}
%This paper specifies the computing infrastructure used for running experiments (hardware and software), including GPU/CPU models; amount of memory; operating system; names and versions of relevant software libraries and frameworks.

All experiments were done on six compute nodes with the following hardwares and softwares. 
\begin{itemize}
\item CPU: AMD Ryzen Threadripper 1950X, 2990WX.
\item GPU: NVIDIA GeForce GTX 1080 Ti, 2 GPUs per node.
\item Memory: 128 GB per node.
\item OS: Linux-5.4.0-122-generic-x86\textunderscore 64-with-glibc2.31.
\item Python version: 3.9.7.
\item numpy version: 1.22.3.
\item stable-baselines3 version: 1.1.0 \cite{stable-baselines3}.
\item gym version: 0.23.1 \cite{openai_gym}.
\item gym-minigrid version: 1.0.2 \cite{gym_minigrid}.
\item ProcGen version: 0.10.7 \cite{cobbe2020procgen}.
\end{itemize}

\newpage
\subsection{Detailed Experimental Results in MiniGrid}
\label{sec:detailed-experimental-results-in-minigrid}

\begin{figure*}[!p]
\centering
\includegraphics[width=1.0\textwidth]{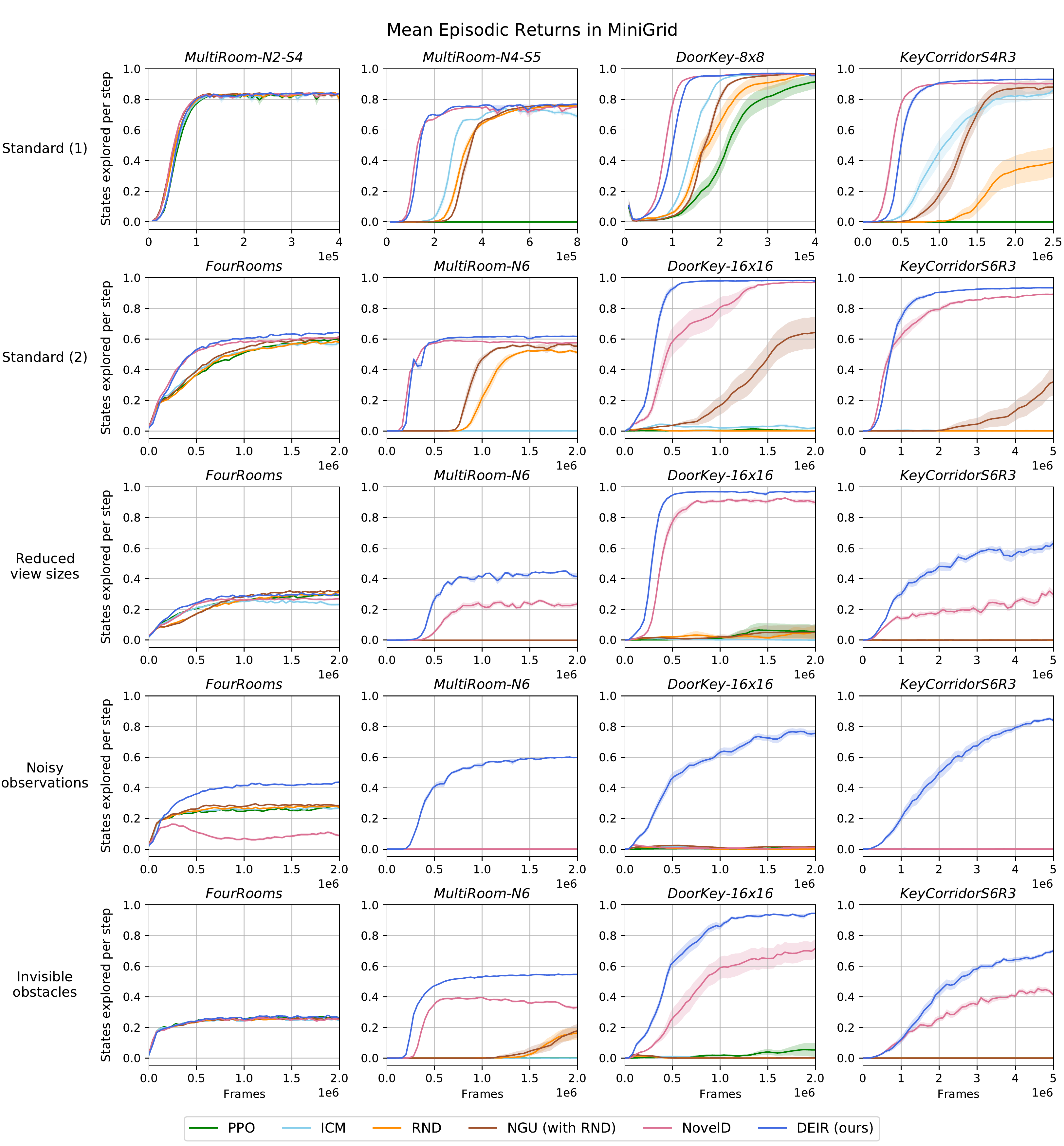}
\caption{Full results of \textbf{mean episodic returns} in standard and advanced MiniGrid games. 
Standard tasks shown in the first row are small in size, consisting of fewer grids in total, and thus are easier than tasks shown in the second row.} 
\label{figFullMinigridReturn}
\end{figure*}

\begin{figure*}[!p]
\centering
\includegraphics[width=1.0\textwidth]{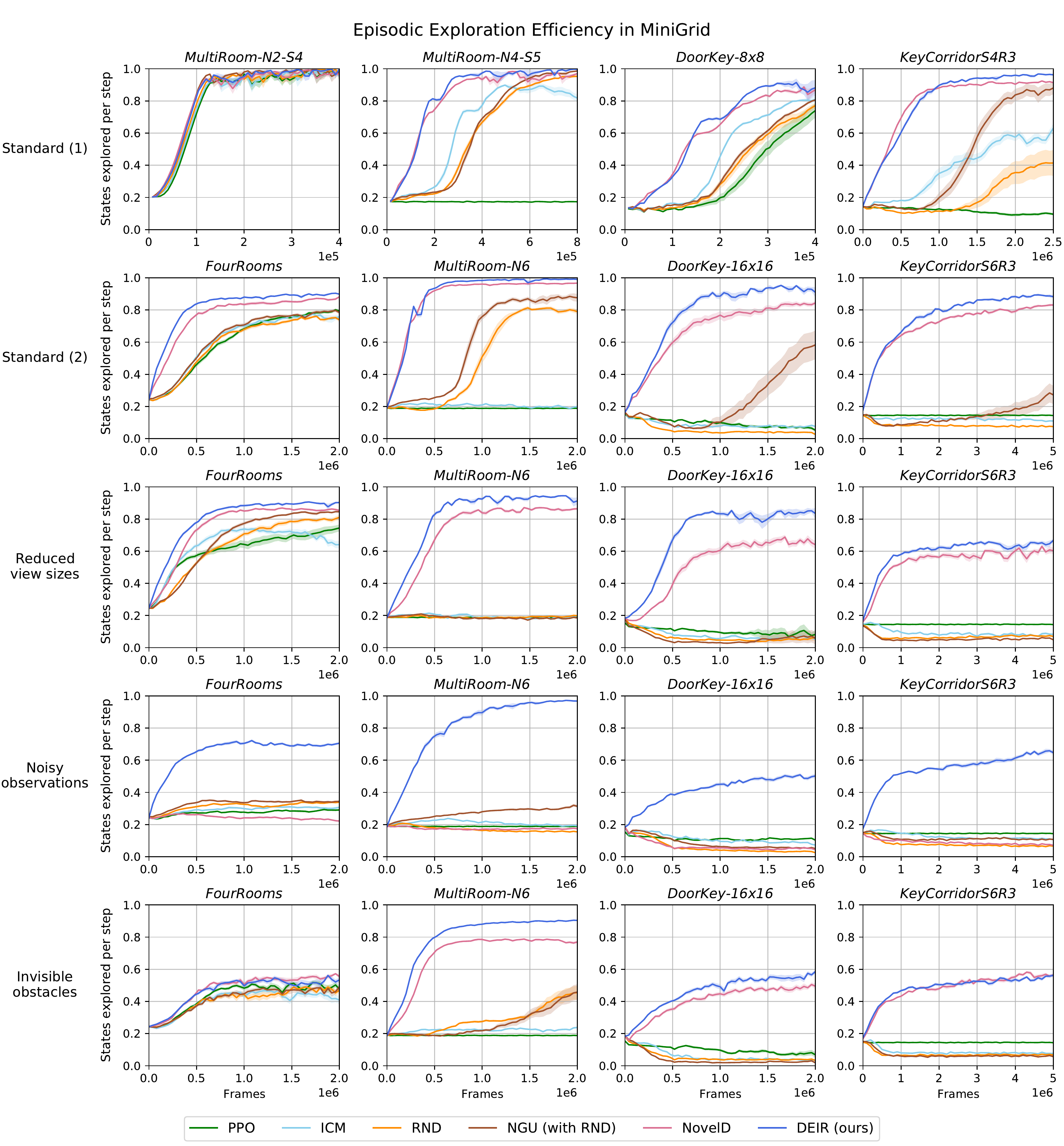}
\caption{Full results of \textbf{episodic exploration efficiency} in standard and advanced MiniGrid games.
Standard tasks shown in the first row are small in size, consisting of fewer grids in total, and thus are easier than tasks shown in the second row.} 
\label{figFullMinigridEpExpl}
\end{figure*}

\begin{figure*}[!p]
\centering
\includegraphics[width=1.0\textwidth]{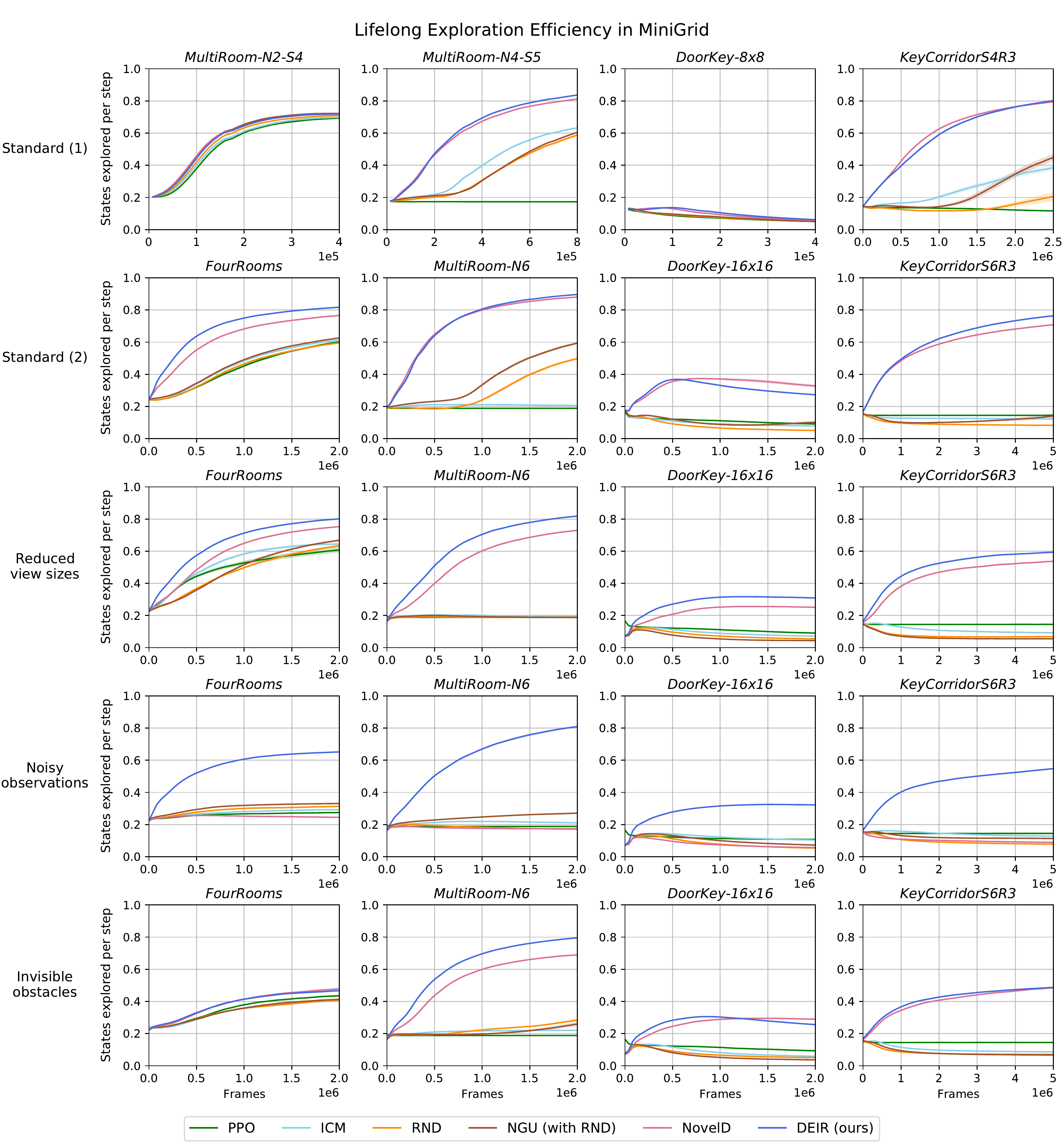}
\caption{Full results of \textbf{lifelong exploration efficiency} in standard and advanced MiniGrid games.
Standard tasks shown in the first row are small in size, consisting of fewer grids in total, and thus are easier than tasks shown in the second row.} 
\label{figFullMinigridLlExpl}
\end{figure*}

In addition to the experimental results reported in the main paper, we show the full  results in MiniGrid games with Figures~\ref{figFullMinigridReturn}, \ref{figFullMinigridEpExpl} and \ref{figFullMinigridLlExpl}.
Particularly, besides mean episodic returns as the main performance metric, we used the below two metrics for evaluating the exploration efficiency of each method.
\begin{itemize}
\item \textbf{Episodic exploration efficiency}: the number of states explored per timestep that have never been seen in the current episode.
\item \textbf{Lifelong exploration efficiency}: the number of states explored per timestep that have never been seen across all previous episodes.
\end{itemize}
Full state information is directly acquired from the backend of MiniGrid games for evaluation and analysis only.
We noticed that the values of states are not equal. An agent may only explore a smaller number of unique states, while those states are actually more valuable overall. However, considering that there is no ground truth available for measuring the absolute value of a state, the two metrics defined above can be used to estimate the exploration efficiency of a method in an objective and relatively accurate way.

The experimental results on exploration efficiency in Figures~\ref{figFullMinigridEpExpl} and \ref{figFullMinigridLlExpl} further clearly suggest that the proposed DEIR can explore more efficiently and robustly than other methods in both standard and advanced MiniGrid environments.

\end{document}